\documentclass{article} 
\usepackage{iclr2024_conference,times}

\usepackage{amsmath,amsfonts,bm}

\def\eqref#1{equation~\ref{#1}}

\def\1{\bm{1}}

\DeclareMathAlphabet{\mathsfit}{\encodingdefault}{\sfdefault}{m}{sl}
\SetMathAlphabet{\mathsfit}{bold}{\encodingdefault}{\sfdefault}{bx}{n}

\usepackage{hyperref}
\usepackage{booktabs} 
\usepackage{url}            
\usepackage{booktabs}       
\usepackage{amsfonts}       
\usepackage{nicefrac}       
\usepackage{microtype}      
\usepackage{xcolor}         
\usepackage{xspace}
\usepackage{soul}
\usepackage{graphicx}
\usepackage{amsmath}
\usepackage{amssymb}
\usepackage{url}
\usepackage{amsthm}
\usepackage{amsfonts}    
\usepackage{nicefrac}       
\usepackage{microtype}      
\usepackage{xcolor}
\usepackage{grffile}
\usepackage{verbatim}
\usepackage{setspace}
\usepackage{multirow}
\usepackage[format=plain,labelformat=simple,labelsep=period,font=small,skip=4pt,compatibility=false,justification=justified]{caption}
\usepackage{booktabs,tabularx}
\usepackage[inline]{enumitem}
\usepackage{mathtools}
\usepackage[ruled,linesnumbered]{algorithm2e}
\usepackage{etoolbox,siunitx}
\robustify\bfseries
\robustify\itshape
\usepackage{newtxtt}
\usepackage{appendix}
\usepackage{wrapfig}
\usepackage[capitalize]{cleveref}
\usepackage{algpseudocode}
\makeatletter
\DeclareRobustCommand\onedot{\futurelet\@let@token\@onedot}
\def\@onedot{\ifx\@let@token.\else.\null\fi\xspace}

\def\eg{\emph{e.g}\onedot} 
\def\ie{\emph{i.e}\onedot} 
\def\cf{\emph{c.f}\onedot}

\makeatother

\newcommand{\myparagraph}[1]{\vspace{-3 pt}\noindent\textbf{#1}}
\newcommand{\model}{\textsc{LRR}} 

\usepackage{pifont}

\definecolor{amber}{rgb}{1.0, 0.75, 0.0}
\newcommand{\blue}[1]{#1}

\title{Look, Remember and Reason: Grounded reasoning in videos with language models}

\author{Apratim Bhattacharyya, Sunny Panchal, Mingu Lee, Reza Pourreza, Pulkit Madan,\\
\textbf{Roland Memisevic} \\
Qualcomm AI Research\thanks{Qualcomm AI Research is an initiative of Qualcomm Technologies, Inc.}
}

\iclrfinalcopy 
\begin{document}

\maketitle

\begin{abstract}
Multi-modal language models (LM) have recently shown promising performance 
in high-level reasoning tasks on videos.
However, existing methods still fall short in tasks like causal or  compositional spatiotemporal reasoning over actions, in which model predictions need to be 
grounded in fine-grained low-level details, such as object motions and object interactions.
In this work, we propose training an LM end-to-end on low-level surrogate tasks, including object detection, re-identification, and tracking, to endow the 
model with the required low-level visual capabilities. 
We show that a two-stream video encoder with spatiotemporal attention is 
effective at capturing the required static and motion-based cues in the video. 
By leveraging the LM's ability to perform the low-level surrogate tasks, 
we can cast reasoning in videos as the three-step process of 
\emph{Look, Remember, Reason}, wherein visual information is extracted using low-level visual skills step-by-step and then integrated to arrive at a final answer. 
We demonstrate the effectiveness of our framework on diverse visual reasoning tasks from the ACRE, CATER, Something-Else and STAR datasets. Our approach is 
trainable end-to-end and surpasses state-of-the-art task-specific methods across tasks by a large margin. 
\end{abstract}

\section{Introduction}
\begin{figure}[t]
  \vspace{-18pt}
  \begin{center}
    \includegraphics[width=0.99\textwidth]{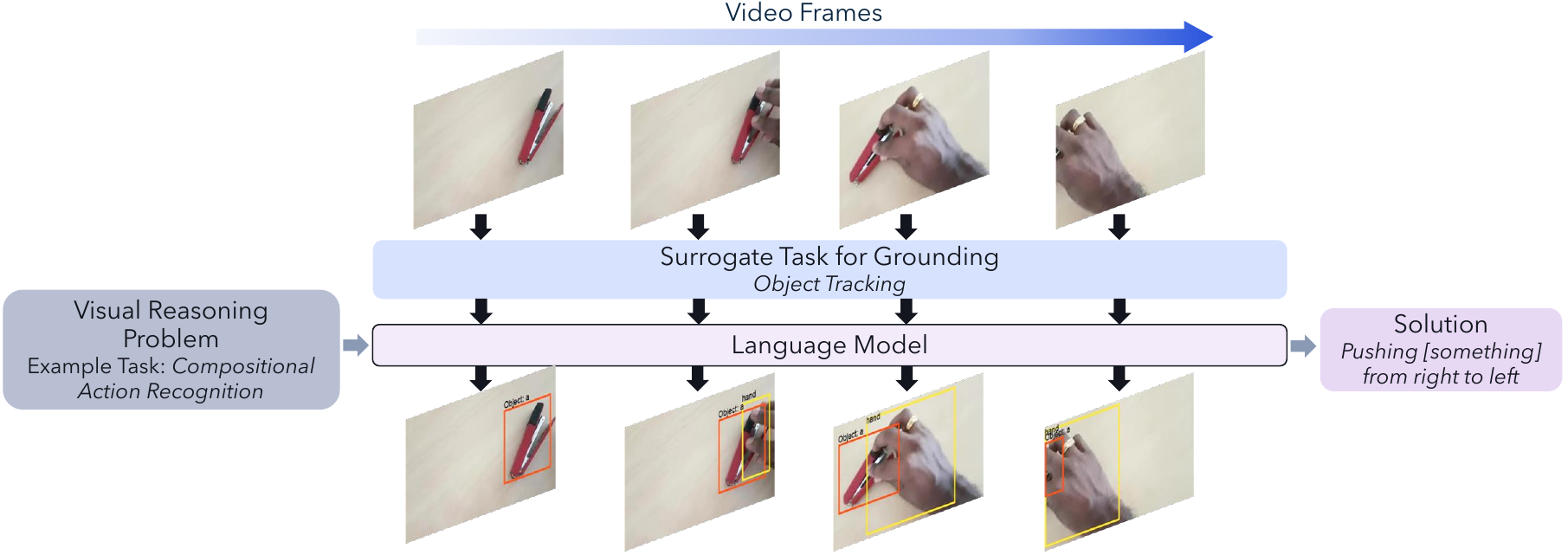}
  \end{center}
  \vspace{-0.2cm}
  \caption{Our Look, Remember, Reason (\model{}) model `looks' at the video frames to extract relevant low-level information, \eg, object motion and interactions, supervised with surrogate tasks like object tracking only during training. It `remembers' the information from intermediate steps and `reasons' using the aggregated information.}
  \label{fig:teaser} 
   \vspace{-0.2cm}
\end{figure}

Autoregressive language models (LMs) have shown impressive results on reasoning tasks such as on grade school math problems \citep{abs-2110-14168} and even on LSAT \citep{abs-2303-08774}. 
Most language models designed for these problems, however, are trained  
on only textual data. 
Since many real-world scenarios require reasoning over heterogeneous 
sensory inputs, \eg, visual cues. 
multi-modal LMs for images or videos have recently gained traction \citep{abs-2204-14198,abs-2301-13823,abs-2302-00923,abs-2306-05424,abs-2305-06355,abs-2305-06355,abs-2306-05424}.
The focus of current multi-modal LMs for videos have primarily been on high-level question answering and instruction following \citep{abs-2306-05424,abs-2305-06355,abs-2204-14198}. 
However, many visual reasoning problems in videos require grounding in fine-grained low-level information, \ie, recognizing objects, and understanding their spatiotemporal interactions. 
For example, as shown in \cref{fig:teaser}, \blue{recognition of the high-level compositional action requires a fine-grained understanding of low-level details of motion and interactions 
between objects}. 
The ability of multi-modal LMs to perform visual reasoning tasks such as compositional action recognition in videos \citep{MaterzynskaXHXW20} that require a combination of low-level skills with high-level reasoning has not yet been explored.

To enable multi-modal LMs to solve such reasoning problems we propose our \emph{Look, Remember and Reason (LRR)} multi-modal LM. 
Our \model{} model architecture extracts dense spatiotemporal features from each input frame. 
This is accomplished using a two-stream attention-based architecture that captures low-level spatial and temporal details \citep{SimonyanZ14} at each input video frame using top-down cross-attention.
Our multi-modal \model{} model is grounded to relevant low-level visual information in the scene by stochastically introducing low-level surrogate tasks during training, including object recognition, re-identification, and tracking, at randomly selected time-steps (\cf \cref{fig:teaser}). 
We keep these grounded spatiotemporal features, which include low-level visual details, in the working memory, \ie., \blue{``remembered'' within the context window of the LM}. 
This allows the model to combine the low-level visual features with high-level inferences to \blue{``reason''} and generate the final responses \blue{using our \model{} framework as shown in \cref{fig:teaser}}. 

Our main contributions are:
\begin{enumerate*}[label=(\roman*)]
\item We highlight the importance of grounding for visual reasoning in multi-modal LMs and propose a novel Look, Remember, and Reason framework to this end to instill the required low-level visual 
skills in the model using surrogate tasks during training; 
\item We introduce a two-stream video encoder that captures both the scene structure and object motions, crucial for learning the low-level skills; 
\item  We demonstrate the effectiveness of our approach on ACRE \citep{0017JEZZ21}, CATER \citep{GirdharR20}, and the real-world Something-Else \citep{MaterzynskaXHXW20} and STAR \citep{WuYC0G21} datasets. Our approach outperforms the prior state-of-the-art, that is based on highly task-specific architectures, by a large margin---highlighting how our general-purpose \model{} model can perform varied and complex spatiotemporal reasoning tasks in videos including causal, compositional and situated reasoning.
\end{enumerate*}

\section{Related Work}

\myparagraph{Multi-modal language models.} 
The use of auto-regressive language models with  adapters to process visual inputs is an active area of research. 
Recently proposed models include Pix2seq \citep{0007SLLFH22}, which is trained to extract low-level visual information from images. Unlike 
our work, extracting low-level visual information is a goal in itself in that work not a surrogate task, and the model processes images not videos. 
Other image-based multi-modal LMs include 
ViperGPT \citep{abs-2303-08128}, VisProg \citep{abs-2211-11559}, and Chameleon \citep{abs-2304-09842}. 
Similarly, PaLM-E \citep{abs-2303-03378} provides images and text as interleaved multi-modal latent vectors, allowing it to process multiple images within any part of a sentence which serves as an input to the LM where the model is trained end-to-end.
LLaMA-Adapter \citep{zhang2023llamaadapter}  introduces an adapter layer to enable multi-modal inputs with the LLaMA model \cite{abs-2302-13971}.
FROMAGe \citep{abs-2301-13823} freezes the language model and fine-tunes the input and output linear layers to encode multi-modal interactions.
InstructBLIP \citep{abs-2305-06500} improves high-level instruction following abilities using instruction fine-tuning.
Qwen-VL \citep{abs-2308-12966} and Kosmos-2 \citep{abs-2306-14824} propose grounded multi-model LMs that can localize referring expressions in images.
In contrast to our work, all these approaches process images not videos. 

Video-based multi-modal LMs include 
Video-ChatGPT \citep{abs-2306-05424}, VideoChat \citep{abs-2305-06355}, and Valley \citep{abs-2306-07207}. They focus on question answering and instruction following using video. 
Flamingo \citep{abs-2204-14198}, by using a
Perceiver as input, is also able to ingest 
video data, and similarly is trained to infer 
high-level concepts. 
In our work, in contrast, we study visual reasoning tasks that require low-level understanding in the form of \eg, motion, actions, and interactions, in addition to high-level reasoning. We show how grounding high-level concepts in these low-level representations 
greatly improves video-based reasoning tasks.

\myparagraph{Spatiotemporal video grounding.} Recent work on grounding focuses on spatially localizing an object in an image or in a video given a referring expression \citep{HuangBDGFN18,0005XGX19,VasudevanDG18,LiWZMZZJW23}. \cite{DengYCZL21,YangC0L20,LuoZSCWDJ20,YangGWHYL19,KamathSLSMC21} propose end-to-end approaches for the spatial localization in images.
For video,  \cite{ChenCMJC18,Chen0CJL19,ZengXHCTG20,ZhangDWWD19,ZhangSJZ20,HeZHLLW19,abs-2308-06947} focus on temporal localization given a natural language query. Methods like STVGBert \citep{SuY021} and STCAT \citep{JinLYM22} perform spatiotemporal localization of a natural language query using a transformer-based model.
\cite{YangMSLS22} focuses on action localization given a natural language query. \cite{abs-2309-01327} propose video grounding using Gaussian mask optimization. 
In contrast to these methods, the approach we introduce performs visual grounding in a 
pre-trained LM through end-to-end training 
using surrogate tasks. 
We show that this makes it possible 
to instill low-level visual capabilities 
as required for solving reasoning problems. 

\myparagraph{Attention-based models and visual reasoning.}
Attention-based models have been studied extensively for visual reasoning \cite{abs-2012-08508,HuARDS17,HudsonM18,KamathSLSMC21,Mahajan020,SantoroRBMPBL17}. Recent advances include an object-centric encoder and a transformer reasoning module to solve RPM-like benchmarks \citep{abs-2303-02260}, multi-hop feature modulation \citep{StrubSPVMPCP18} and cascaded modulation networks \citep{YaoXWX18} that use a multi-step comprehension process, neural interpreters \citep{RahamanGJGBLS21} that factorize inference in a self-attention network and ALANS learner \citep{ZhangXJWZZ22} that combines abstract algebra and representation theory. Calibrating concepts and operations \cite{LiSZXTDY21} enables neural symbolic models to capture underlying data characteristics and perform hierarchical inference.
Again, the key difference in our work is that we instill the ability to extract object-centric information within the LM itself, using surrogate tasks, instead of resorting to separate object detection modules. 

\section{Look, Remember, Reason}
\blue{To allow for visual reasoning in videos using language models, we propose a novel \emph{Look, Remember, Reason} framework which grounds the model to the low-level structual and motion information in the visual input required for solving high-level reasoning tasks. Grounding is enabled through low-level visual surrogate tasks, \eg object recognition, tracking and re-identification.
In the following, we first describe our auto-regressive \model{} architecture including details of our two-stream video encoder, followed by our training pipeline with details of our surrogate tasks}.

\subsection{Auto-regressive Pipeline}
\begin{figure*}[t]
  \centering
  \includegraphics[width=0.975\linewidth]{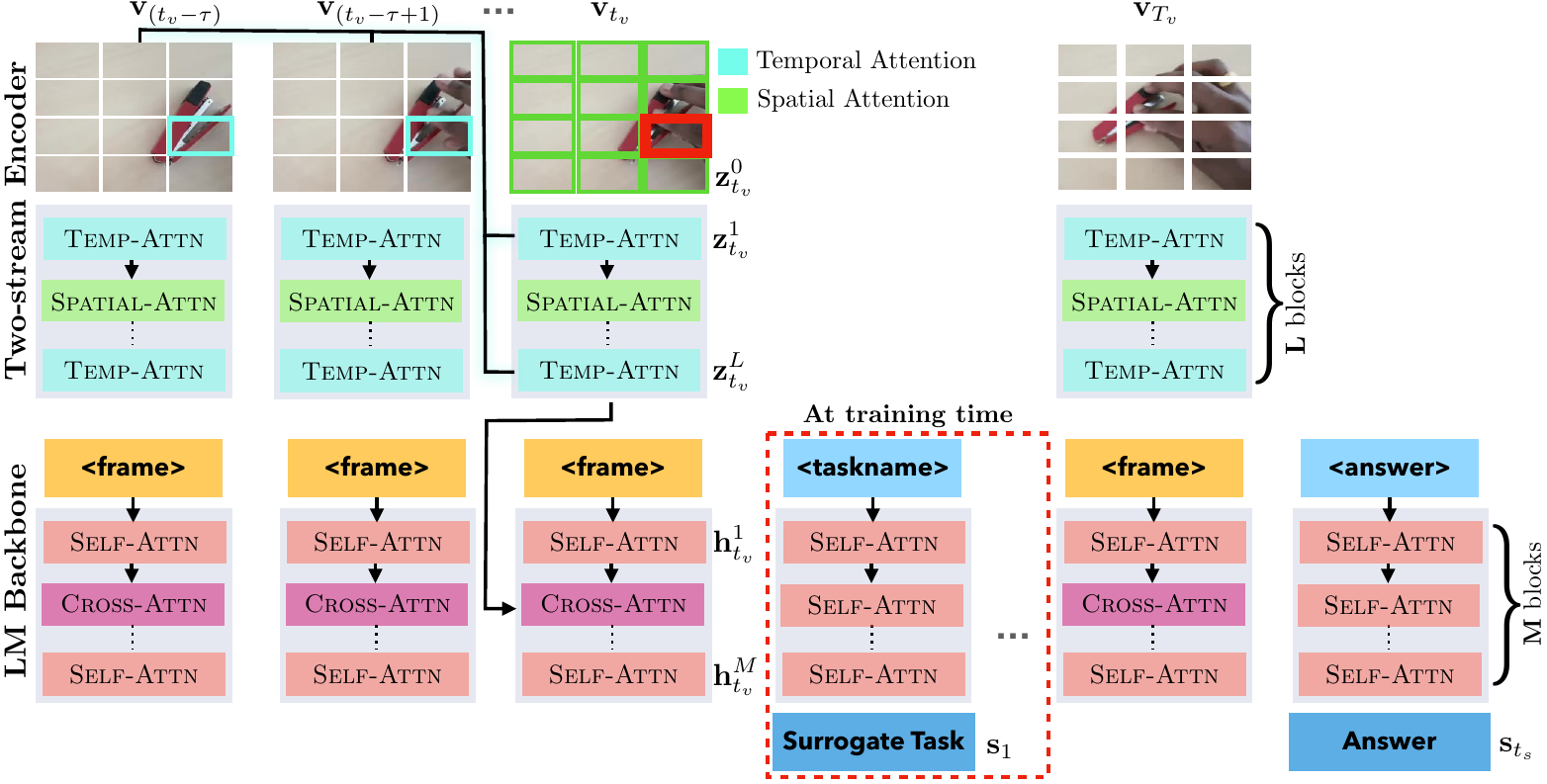}
\caption{The architecture of our \model{} model, highlighting the use of interleaved top-down cross-attention layers in between self-attention layers higher up in the hierarchy. }
\label{fig:architecture} 
\vspace{-0.5cm}
\end{figure*}

\blue{Our \model{} model parameterized by $\theta$, as shown in \cref{fig:architecture}, is based on a pre-trained LM backbone with a two-stream auto-regressive video encoder. Our \model{} model receives the visual reasoning problem $\mathbf{Q}$ and, an interleaved sequence $\mathcal{I}$ of video frames $\mathbf{V} = ( \mathbf{v}_1, \dots, \mathbf{v}_{T_v} )$ and tokenized text $\mathbf{S} = ( \mathbf{s}_1, \dots, \mathbf{s}_{T_s} )$ as input. $\mathbf{S}$ consists of low-level visual surrogate tasks and the answer indicated by <taskname> and <answer> respectively.}
The input video frame sequence $\mathbf{V}$ is encoded by our two-stream video encoder.
\blue{The LM backbone receives as input $\mathbf{Q}$, and the interleaved sequence of encoded video frames $\mathbf{V}$ whose positions are indicated with <frame> special tokens and $\mathbf{S}$.}
Conditioned on $\mathbf{Q}$ and $\mathbf{V}$, we train our \model{} model by maximizing log-likelihood of the text sequence $\mathbf{S}$ as,
\begin{align}\label{eq:autopipeline}
    \log\big( p_\theta(\mathbf{S} | \mathbf{Q}, \mathbf{V} )\big) = \sum_{t_s} \log\big( p_\theta(\mathbf{s}_{t_s} | \mathbf{s}_1, \dots, \mathbf{s}_{t_s - 1}, \mathbf{v}_1, \dots, \mathbf{v}_{t_v}, \mathbf{Q} ) \big)
\end{align}
where, $(\mathbf{v}_1, \dots, \mathbf{v}_{t_v})$ is the interleaved video frame sub-sequence up to the text token $\mathbf{s}_{t_s}$. 
In \cref{sec:surrogate}, we describe the surrogate tasks included in $\mathbf{S}$.
The parameters of the LM backbone are initialized from pre-trained LMs, allowing us to exploit their existing high-level reasoning capabilities. 
We use LMs from the OPT family \citep{abs-2205-01068}, but verified that similar performance can be achieved using other pre-trained models \citep{abs-2101-00027,abs-2211-05100}. 
While LM backbones are trained only on text, visual reasoning relies on information from the visual domain $\mathbf{V}$. Therefore, to ground our \model{} model, visual information in $\mathbf{V}$ needs to be mapped to the text-based representation space of the LM. The key challenge here is that in contrast to text tokens, videos are highly information dense.
To address this, we develop a two-stream video encoder to extract information relevant for the visual reasoning problem at hand, such that responses to the visual reasoning problems are grounded in the fine-grained low-level details of the video.

\subsection{Two-stream video encoder}
Our autoregressive two-stream video encoder exploits divided space-time attention \citep{BertasiusWT21} and  \blue{generates for each input video frame patch based embeddings that capture relevant low-level spatio-temporal information. Spatial attention captures structural information, \eg, object identities. Temporal attention autoregressively captures object motion and interaction information using the previous $\tau$ frames as a buffer}.
As a first step, it converts each input video frame $\mathbf{v}_{t_v}$ into $P$ flattened patches $\mathbf{v}_{t_v} = \{ \mathbf{v}_{(1, t_v)}, \dots, \mathbf{v}_{(P, t_v)}  \}$ of size $16\times16$ \citep{DosovitskiyB0WZ21} as shown in \cref{fig:architecture}. Our \model{} model applies a linear transformation ($\textsc{Ln}$) on the input patches to generate the initial patch embeddings $\mathbf{z}^0_{t_v} = \{ \mathbf{z}^0_{(1, t_v)}, \dots, \mathbf{z}^0_{(P, t_v)}  \}$ as,
\begin{align}
    \mathbf{z}^0_{(p, t_v)} = \textsc{Ln}(\mathbf{v}_{(p, t_v)}) + \mathbf{em}_{(p, t_v)}
\end{align}
The spatiotemporal positional embedding $\mathbf{em}_{(p, t_v)}$ is added to each patch to aid our video encoder extract spatiotemporal features from each patch in the input video frame. Next, these patch embeddings are processed by the $L$ blocks of our video encoder where at each block spatial or temporal attention is applied. At every block $\ell \in \{1, \dots, L \}$ of our model, the embeddings are linearly mapped to produce the query, keys, and values required for the space- and time-based attention operations,
\begin{align}
    \mathbf{q}^\ell_{(p, t_v)} = \textsc{Ln}(\mathbf{z}^\ell_{(p, t_v)}), \,\,\, \mathbf{k}^\ell_{(p, t_v)} = \textsc{Ln}(\mathbf{z}^\ell_{(p, t_v)}), \,\,\, \mathbf{v}^\ell_{(p, t_v)} = \textsc{Ln}(\mathbf{z}^\ell_{(p, t_v)}).
\end{align}
In case of temporal attention, for the patch embedding $\mathbf{z}^\ell_{(p, t_v)}$ the attention is computed over patch embeddings at the same spatial position $p$ in the $\tau$ previous input video frames, $\mathbf{z}^\ell_{(p, (t_v-\tau))}$ to $\mathbf{z}^\ell_{(p, t_v)}$. For the patch highlighted in red in \cref{fig:architecture} the patches where temporal attention is applied in the previous $\tau$ input frames is highlighted in {\setulcolor{cyan} \ul{cyan}}. The temporal attention vector $(\mathbf{\alpha}_\mathcal{T})^\ell_{(p, t_v)}$ is given by,
\begin{align}
    (\mathbf{\alpha}_\mathcal{T})^\ell_{(p, t_v)} = \textsc{Softmax}\Big( \frac{\mathbf{q}^\ell_{(p, t_v)}\phantom{}^\top}{\sqrt{d_m}} [ \mathbf{k}^\ell_{(p, (t_v-\tau))}, \dots, \mathbf{k}^\ell_{(p, t_v)} ] \Big)
\end{align}
where $d_m$ is the dimensionality of the key, queries and values. In contrast, the spatial attention vector, $(\mathbf{\alpha}_\mathcal{S})^\ell_{(p, t_v)}$, is calculated over all patch embeddings of the current input video frame, highlighted in {\setulcolor{green} \ul{green}} in \cref{fig:architecture},
\begin{align}
    (\mathbf{\alpha}_\mathcal{S})^\ell_{(p, t_v)} = \textsc{Softmax}\Big( \frac{\mathbf{q}^\ell_{(p, t_v)}\phantom{}^\top}{\sqrt{d_m}} [ \mathbf{k}^\ell_{(1, t_v)}, \dots, \mathbf{k}^\ell_{(P, t_v)} ] \Big)
\end{align}
The patch embeddings $\mathbf{z}^{\ell+1}_{(p, t_v)}$ are the weighted sums of the values $\mathbf{v}^\ell_{(p, t_v)}$ computed using the attention weights $(\mathbf{\alpha}_\mathcal{T})^\ell_{(p, t_v)}$ or $(\mathbf{\alpha}_\mathcal{S})^\ell_{(p, t_v)}$ in case of temporal or spatial attention, respectively. Finally, the patch embeddings ${\mathbf{z}}^L_{t_v}$, containing localized information of the scene structure and motion, are obtained as output from the last block $L$ at time-step $t_v$ of the video encoder.

\subsection{LM Backbone with Top-down Cross Attention}
\blue{Visual information encoded by the two-stream video encoder is mapped top-down to the LM backbone at positions indicated by the <frame> token at time-steps $t_v$ (\cref{fig:architecture}) using cross attention layers. This top-down mechanism with cross attention layers, trained using surrogate tasks (\cref{sec:surrogate}), helps us ground the LM backbone to the relevant low-level information from the input video frames.}

We modify the backbone LM architecture by inserting cross attention (\textsc{Cross-Attn}) layers \citep{abs-2204-14198} between self-attention (\textsc{Self-Attn}) layers (\cref{fig:architecture}). 
The output patch embeddings from our video encoder $\mathbf{z}^L_{t_v} = \{\mathbf{z}^L_{(1, t_v)}, \dots, \mathbf{z}^L_{(p, t_v)}\}$ are transformed using multi-layer perceptrons (\textsc{Mlp}) at every cross-attention layer which help deal with the domain shift between the visual and textual domains. Next, we exploit the rich hierarchical representations in the hidden states $\mathbf{h}_{t_v} = \{ \mathbf{h}^1_{t_v}, \ldots, \mathbf{h}^M_{t_v}\}$ of the LM at timestep ${t_v}$, where $M$ is the number of self-attention layers, \blue{to ``look'' at the visual input and extract visual information in a top-down fashion for grounding}. Specifically, we use the representation $\hat{\mathbf{h}}^k_{t_v}$ after the application of the $(k + 1)^\text{th}$ self-attention layer in the LM. \blue{The hidden representation $\hat{\mathbf{h}}^k_{t_v}$ encodes global semantics about the visual reasoning problem $\mathbf{Q}$ and previous video frames upto ${t_v}$}. It serves as the query vector after a linear transformation and the visual features $\mathbf{z}^L_{t_v}$ serves as the keys and values of the cross attention layer respectively,
\begin{align}
&\hat{\mathbf{h}}^k_{t_v} = \textsc{Self-Attn}(\mathbf{h}^{k}_{t_v}) \label{eq:xattn:1}\\
&\hat{\mathbf{z}}_{t_v} = \textsc{Cross-Attn}(\hat{\mathbf{h}}^k_{t_v}, \mathbf{z}^L_{t_v})\label{eq:xattn:2}\\
&\mathbf{h}^{k+1}_{t_v} = \mathbf{h}^k_{t_v} + \hat{\mathbf{h}}^k_{t_v} + \hat{\mathbf{z}}_{t_v} \label{eq:xattn:3}\\
&\mathbf{h}^{k+1}_{t_v} = \textsc{FFN}(\mathbf{h}^{k+1}_{t_v}) + \mathbf{h}^{k+1}_{t_v} \label{eq:xattn:4}
\end{align}
where FFN is a feedforward layer \citep{VaswaniSPUJGKP17}. 
\blue{As the hidden representation $\mathbf{h}^k_{t_v}$ encodes global semantics, it allows the LM backbone to extract relevant low-level spatiotemporal information in $\hat{\mathbf{z}}_{t_v}$ (\cref{eq:xattn:2}), when trained using our surrogate tasks.} This information added to  hidden states $\mathbf{h}^{k+1}_{t_v}$ (\cref{eq:xattn:3}) and is implicitly "remembered" within the context window of the LM. \blue{This information can be aggregated by the LM to ``reason'' and arrive at the final answer.}

\subsection{Grounding through Surrogate tasks}
\label{sec:surrogate}
To ground our \model{} model to the relevant low-level information in the visual input, we utilize surrogate tasks during training. This is illustrated in \cref{fig:architecture} where the model is prompted using special tokens (<taskname>) to solve a surrogate task. \blue{We consider tasks like object recognition, tracking and re-identification as low-level. From \cref{fig:teaser}, the ability to recognize the performed action rests upon grounding to the motion and interactions of the hand and the stapler. This makes recognition and tracking constituent low-level capabilities for recognizing the (high-level) compositional action.}

Our \model{} model is highly flexible and can be prompted to solve a wide range of low-level surrogate tasks. Here, we focus primarily on the tasks of object recognition, localization, re-identification and tracking. These tasks are fundamental to solving a range of visual reasoning problems, and the requisite ground truth can be readily obtained using off-the-shelf vision models \citep{RenHG017,TangAAS17,YeSLXSH22}. Moreover, these tasks can be encoded as text to be processed by the LM backbone in the following general format: ``<taskname> \emph{object id$_1$, object class$_1$, object bounding box$_1$; $\dots$ ; object id$_n$, object class$_n$, object bounding box$_n$}'', where there are $n$ objects in the scene. 
The <taskname> special token prompts the model to solve the surrogate task. We use <detect>, <re-identify> and <track> for detection, re-identification, and tracking surrogate tasks respectively. 
The \emph{object id} is an integer that is assigned by the model based on the spatiotemporal order of appearance and is crucial for re-identification and tracking as same object instance should be assigned the same id across video frames. The bounding box is described as a 4-tuple of the x and y coordinates of the upper left and lower right corners. We provide additional details in the experimental section (\cref{sec:experiments}).

Our \model{} model is prompted to solve these surrogate tasks at randomly selected time steps during training.
Random prompting forgoes the need to include surrogate tasks during inference time, leading to faster inference. 
Random prompting also benefits training efficiency in the case of long video sequences. 
To train our \model{} model using maximum likelihood as described in \cref{eq:autopipeline}. We construct the token sequence $\mathbf{S}$ in $\mathcal{I}$ that includes surrogate tasks for each training example as follows: 
    We first randomly select a subset of video frames  $( \mathbf{v}_{t_1}, \dots, \mathbf{v}_{t_k} ) \in \mathbf{V}$.
    We then update the token sequence $\mathbf{S}$ to include the surrogate tasks at these randomly selected time-steps, the beginning of which is marked by a task specific <taskname> special token, \eg, <detect>, <re-identify> and <track>. The surrogate task itself is added in $\mathbf{S}$ as described above.

\section{Experiments}\label{sec:experiments}
\blue{We evaluate on visual reasoning tasks from: ACRE \citep{0017JEZZ21}, Something-Else \cite{MaterzynskaXHXW20}, CATER \citep{GirdharR20} and STAR \citep{WuYC0G21}}.

\myparagraph{Models and training details.} We fine-tune the pre-trained LM backbone along with the video encoder and cross-attention layers in our \model{} model as the visual reasoning problems considered here are challenging and cannot be accurately solved by prompting state of the art LMs such as GPT-4 \citep{abs-2305-19555}.
We focus on the OPT family of LMs \cite{abs-2205-01068}, particularly OPT-125M/350M/1.3B.
We use 4 Nvidia A100 GPUs. Additional details in the appendix.

\begin{table*}[!t]
\scriptsize
\centering
\caption{Evaluation results on the ACRE dataset, where, D.R. -- Direct evidence, I.D. -- Indirect evidence, S.O. -- Screened-off, and B.B. -- Backward Blocked subsets (\blue{$^*$represents results tested by ourselves.}). }
\label{tab:acre_eval}
\vspace{-0.15cm}
\begin{tabularx}{\linewidth}{@{}X|ccccc|ccccc@{}}
\toprule
& \multicolumn{5}{c}{Compositional} & \multicolumn{5}{c}{Systematic} \\
Model & All & D.R. & I.D. & S.O. & B.B. & All & D.R. & I.D. & S.O. & B.B  \\
\midrule
NS-OPT \citep{0017JEZZ21} & 69.0 & 92.5 & 76.0 & 88.3 & 13.4  & 67.4 & 94.7 & 88.3 & 82.7 & 16.0 \\
IV-CL$^{\dag}$ \citep{abs-2307-08506} & 93.2 & - & - & - & - & 92.6 & - & - & - & -\\
ALOE \citep{abs-2012-08508} & 91.7 & 97.1 & {90.8} & {96.8} & 78.8 & 93.9 & 97.1 & 71.2 & {98.9} & 94.4\\
OpenFlamingo$^*$ \citep{abs-2308-01390} & 38.2 & 42.6 & 49.5 & 9.9 & 47.6 & 38.6 & 36.5 & 25.8 & 13.7 & 67.6\\
\midrule
\model{} (Ours) & \textbf{98.2} & \textbf{99.9} & \textbf{92.8} & \textbf{99.2} & \textbf{97.4} & \textbf{99.2} & \textbf{99.9} & \textbf{97.0} & \textbf{99.9} & \textbf{98.8}\\
\midrule
\model{} (w/o Surrogate tasks) & 38.1 & 38.4 & 30.2 & 26.2 & 50.0 & 36.5 & 35.2 & 28.1 & 20.0 & 55.3  \\
\model{} (w/o “Two-stream” encoder) & 92.2 & 98.6 & 77.7 & 96.2 & 85.8 & 92.8 & 98.5 & 76.2 & 96.8 & 89.4 \\
\model{} (from scratch) &  87.7 & 96.9 & 57.7 & 91.1 & 84.8 & 88.0 & 96.3 & 58.5 & 91.5 & 86.9\\
\bottomrule
\end{tabularx}
\end{table*}

\begin{figure*}[!t]
\scriptsize
\centering
\begin{tabularx}{0.955\textwidth}{@{}ccccc|c@{}}
\toprule
\multicolumn{5}{c}{Context Trials} & Query \\ 
\midrule
\includegraphics[width=0.1155\linewidth]{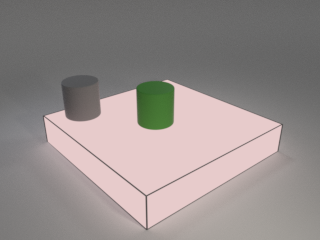} &
 &
\includegraphics[width=0.1155\linewidth]{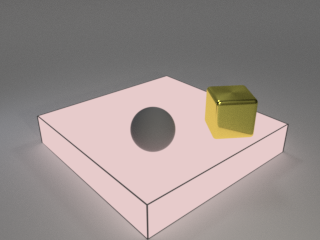} &
 &
\includegraphics[width=0.1155\linewidth]{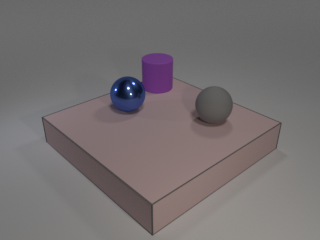} &
\includegraphics[width=0.1155\linewidth]{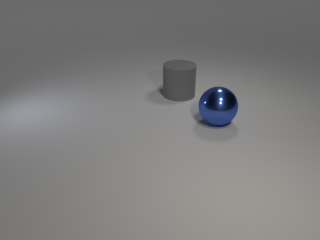} \\

\raggedright
\parbox{2.6cm}{<re-identify> \emph{ {\setulcolor{red} \ul{1,medium gray rubber cylinder}}; 2, medium green rubber cylinder.} \\ <blicket> \emph{on.}} &
$\dots$ &
\parbox{2.6cm}{<re-identify>  \emph{{\setulcolor{blue} \ul{3, medium gray rubber sphere}}; 4, medium yellow metal cube.} \\ <blicket> \emph{on.}} &
$\dots$ &
\parbox{2.6cm}{<re-identify>  \emph{{\setulcolor{green} \ul{5, medium blue metal sphere}}; 6, medium purple rubber cylinder; {\setulcolor{blue} \ul{3, medium gray rubber sphere.}}} \\ <blicket> \emph{off.}} &

\parbox{2.65cm}{<re-identify> \emph{{\setulcolor{red} \ul{1, medium gray rubber cylinder}}; {\setulcolor{green} \ul{5, medium blue metal sphere.}}} \\ <answer> \emph{no.}}
\\
\bottomrule
\end{tabularx}
\vspace{-0.1cm}
\caption{Example solutions to surrogate tasks generated by our \model{} model on ACRE. Re-identified objects across context trials are underlined in the same color. }
\label{tab:acre_qualitative}
\end{figure*}

\subsection{ACRE}\label{sec:acre}
The ACRE dataset \citep{0017JEZZ21} evaluates the performance of vision systems on the problem of causal induction. Specifically, the dataset focuses on the problem of causal discovery using ``blicket'' detection tests, originally administered to children, using a blicket detector which activates when a blicket object is placed on it. The experiment involves a series of context trials, in which various (combinations of) objects are placed on the blicket detector, and subjects are shown whether the detector is activated. They are then asked which objects or (novel) combinations of objects would activate the machine. The ACRE dataset contains 6 context trials per example and 4 ``blicket'' detection tests based on the images.

\myparagraph{Surrogate tasks.} The key low-level visual challenge in the ACRE dataset is to associate query objects to the context trials to detect whether the blicket machine is activated. Therefore, we consider the surrogate tasks of object recognition and re-identification. The solution to the surrogate task can be generated by the backbone LM in the following format: ``<re-identify> \emph{object id$_1$,  object class$_1$; $\dots$ ; object id$_n$, object class$_n$} '' as shown in \cref{tab:acre_qualitative}. We also found it helpful to introduce an additional surrogate task to identify the state of the blicket machine: ``<blicket> \emph{on/off}''. These surrogate tasks are introduced randomly during training with a probability of 30\% after each context trial or query.

\myparagraph{Baselines and evaluation.} We base our \model{} models on the OPT-125M backbone. We compare to the state of the art in \cref{tab:acre_eval}, including the \blue{(powerful transformer based) multi-modal LLM} OpenFlamingo (3B-mosaicml/mpt-1b-redpajama-200b-dolly; \cite{abs-2308-01390}) as it has shown success in reasoning problems involving multiple images and videos. To highlight the importance of our rationale generation process, we consider a baseline without the surrogate re-identification task: \model{} (w/o Surrogate tasks).
Although there is no motion or object interactions in ACRE, we still found it helpful to use our two-stream encoder as our temporal attention mechanism can be used for re-identification of (previously observed) objects. To highlight this, we also include a \model{} (w/o Two-stream encoder) ablation that uses a plain single stream ViT encoder.
Finally, to highlight the importance of using pre-trained LM backbones we include a \model{} model that is trained from scratch with the same OPT-125M LM architecture.

Our \model{} model outperforms the \blue{state of the art powerful transformer based ALOE \citep{abs-2012-08508}, by a large margin of 6.5\% and 5.3\% on the compositional and systematic splits respectively. 
This shows the advantage of our end-to-end \model{} model with surrogate tasks over explicit object centric input representations used by ALOE}.
Similarly, we observe a significant performance gain of 60\% over the \blue{powerful transformer based multi-modal LLM} OpenFlamingo \citep{abs-2308-01390} which highlights the importance of our \model{} framework to ground our model to the relevant low-level details -- highlighted by the weak performance of the LRR (w/o Surrogate tasks) ablation. 
We also see that our two-stream encoder improves performance over the single-stream model.
The weak performance of the \model{} (from scratch) ablation shows that it is crucial to start from a pre-trained LM backbone to exploit its high level reasoning abilities. 
Finally, although the results in \cref{tab:acre_eval} use the OPT-125M backbone in our \model{} model, we obtain an identical 98.2\% and 99.0\% accuracy on the compositional and systematic splits respectively with an OPT-1.3B backbone, indicating that our \emph{look, remember, reason} framework is applicable across LM backbone sizes. 

\begin{table*}[t]
\scriptsize
\centering
\caption{Evaluation on the Something-Else dataset (\blue{$^*$represents results tested by ourselves.}). } 
\vspace{-0.15cm}
\label{tab:somethingelse_eval}
\begin{tabularx}{\linewidth}{@{}X|cc|cc@{}}
\toprule
& \multicolumn{2}{c}{Base} & \multicolumn{2}{c}{Compositional} \\
Method & Top-1 & Top-5 & Top-1 & Top-5 \\
\midrule
STIN + OIE + NL \citep{MaterzynskaXHXW20} & 78.1 & 94.5 & 56.2 & 81.3 \\
Video-ChatGPT$^*$ \citep{abs-2306-05424} & 52.6 & 75.8 & 38.6 & 67.8\\
\midrule
\model{} (Ours) & \textbf{80.2} & \textbf{96.1} & \textbf{62.0} & \textbf{86.3}\\
\midrule
\model{} (w/o Surrogate tasks) & 71.3 & 89.6 & 50.1 & 70.8 \\
\model{} (w/o Two-stream encoder) & 73.2 & 90.4 & 53.6 & 76.1  \\
\bottomrule
\end{tabularx}
\end{table*}

\begin{figure*}[t]
\small
\centering
\begin{tabular}{ccc}

\includegraphics[width=0.29\linewidth]{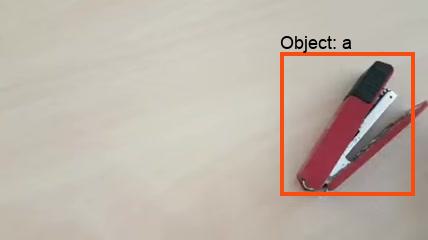} & 
\includegraphics[width=0.29\linewidth]{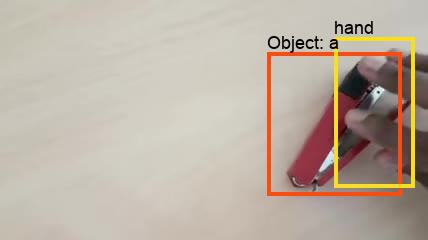} &
\includegraphics[width=0.29\linewidth]{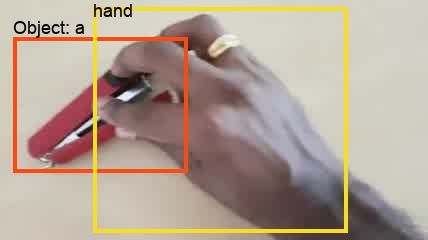} \\

\includegraphics[width=0.29\linewidth]{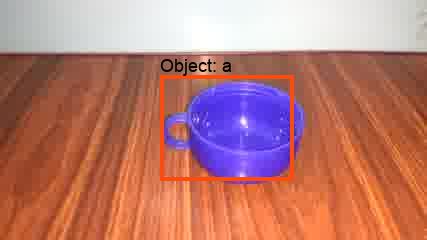} & 
\includegraphics[width=0.29\linewidth]{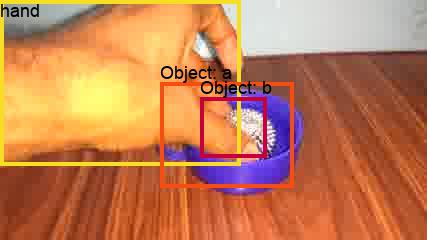} &
\includegraphics[width=0.29\linewidth]{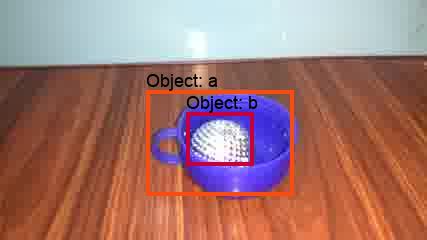} \\

\end{tabular}
\vspace{-0.1cm}
\caption{Example solutions to surrogate task tracking generated by our \model{} model on Something-Else. Bounding boxes belonging to the same track are highlighted using the same color.}
\label{tab:somethingelse_qualitative}
\vspace{-0.5cm}
\end{figure*}

\subsection{Something-Else}\label{sec:somethingelse}
The \blue{complex real-world} Something-Else dataset \citep{MaterzynskaXHXW20} focuses on compositional action recognition. It is based on the Something-Something dataset \citep{GoyalKMMWKHFYMH17}, which it extends to measure compositional generalization. The compositional split of the dataset breaks each action down into a combination of a verb, a subject, and one or more objects. 
This makes 
it possible to benchmark performance on novel verb and object combinations which are not present in the training set, requiring models to have a fine-grained understanding of motion instead of merely correlating actions with object types.

\myparagraph{Surrogate tasks.} 
In this task, we employ tracking as a surrogate task to support 
the model's ability to capture motion and object interactions. 
Note that, as object classes are not important for compositional action recognition, the solution to the surrogate task can be generated in the following format: ``<track> \emph{object id$_1$,  object bounding box$_1$; $\dots$ ; object id$_n$, object bounding box$_n$}'' as shown in \cref{tab:somethingelse_qualitative}.
Surrogate tasks are introduced randomly during training with a probability of 30\% after each input video frame. 

\myparagraph{Baselines and evaluation.} We base our \model{} model on the OPT-125M LM backbone and compare to several baselines and ablations in \cref{tab:somethingelse_eval}. We report results on both the base split and the compositional split with novel action-object combinations. We consider the state-of-the-art STIN + OIE + NL models from \cite{MaterzynskaXHXW20} along with the \blue{state-of-the-art powerful transformer based multi-modal LLM} Video-ChatGPT \citep{abs-2306-05424} model as baselines. The Video-ChatGPT model uses the 7B parameter Vicuna model \citep{vicuna2023} as an LM backbone and a CLIP \citep{RadfordKHRGASAM21} based video encoder.
To highlight the importance of our surrogate tasks, we consider an ablation without the surrogate tracking task: \model{} (w/o Surrogate Tasks).
To highlight the importance of our two-stream video encoder we consider an ablation with a single stream ViT model that cannot capture motion features (w/o Two-stream encoder).

Our \model{} model also outperforms the STIN + OIE + NL baseline by 2.1\% and 5.8\% Top-1 accuracy on the base and compositional split respectively, which highlights the reasoning ability of grounded LM based architectures.
The results also show that the performance of the \blue{state of the art multi-modal LM}: Video-ChatGPT (\blue{finetuned}), lags very significantly by 23.4\% behind our \model{} model. This is because of the CLIP based video encoder in Video-ChatGPT is not well suited for capturing motion features and due to a lack of grounding. The importance of motion features is underscored by the lacking performance of the plain ViT based (w/o Two-stream encoder) ablation: a Top-1 accuracy drop of 8.4\% on the compositional split. The importance of grounding is highlighted through our ablation without surrogate tasks: a Top-1 accuracy drop of 11.9\% on the compositional split. We illustrate successful tracking of objects \blue{in complex real-world scenarios} by our \model{} model in \cref{tab:somethingelse_qualitative} \blue{and in Appendix B}. Note that, although the results in \cref{tab:somethingelse_eval} uses the OPT-125M backbone, we obtain an identical 61.3\% Top-1 and 85.9\% Top-5 accuracy on the compositional split with an OPT-1.3B backbone, indicating that our \emph{look, remember, reason} framework is applicable across LM sizes. 

\begin{table*}[t]
\scriptsize
\centering
\caption{Evaluation on the CATER dataset ($^{\dag}$results reported only for static camera). }
\vspace{-0.15cm}
\label{tab:cater_eval}
\begin{tabularx}{\linewidth}{@{}X|ccc|ccc@{}}
\toprule
& \multicolumn{3}{c}{Static Camera} & \multicolumn{3}{c}{Moving Camera} \\
 Method &  Top-1($\uparrow$) & Top-5($\uparrow$) & L1(grid;$\downarrow$) & Top-1($\uparrow$) & Top-5($\uparrow$) & L1(grid;$\downarrow$)  \\
\midrule
R3D + NL LSTM \cite{GirdharR20} & 46.2 & 69.9 & 1.5 & 38.6 & 70.2 & 1.5 \\
ALOE \cite{abs-2012-08508} & 74.0 & 94.0 & 0.44 & 59.7 & 90.1 & 0.69\\
\midrule
IV-CL$^{\dag}$ \citep{abs-2307-08506} & 70.1 & 88.3 & - & - & - & -\\
OPNet$^{\dag}$ \citep{abs-2003-10469} & 74.8 & - & 0.54 & - & - & - \\
Hopper$^{\dag}$ \citep{ZhouKLNMKG21} & 73.2 & 93.8 & 0.85 & - & - & - \\
TFC V3D$^{\dag}$ \citep{abs-2203-05928} & 79.7 & 95.5 & 0.47 & - & - & - \\
\model{} (Ours) & \textbf{84.1} & \textbf{97.2} & \textbf{0.34} & \textbf{80.4} & \textbf{96.7} & \textbf{0.42}\\
\midrule
\model{} (w/o Surrogate tasks) & 68.5 & 88.7 & 0.65 & 62.7 & 86.7 & 0.77\\
\model{} (w/o Two-stream encoder) & 81.4 & 97.2 & 0.44 & 75.6 & 96.6 & 0.53\\
\bottomrule
\end{tabularx}
\end{table*}

\begin{figure*}[t]
\small
\centering
\begin{tabularx}{0.88\linewidth}{@{}cccc@{}}

\includegraphics[width=0.2\linewidth]{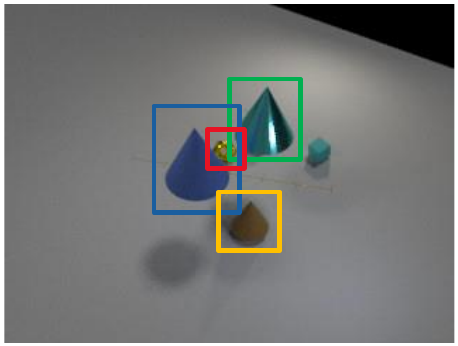} & 
\includegraphics[width=0.2\linewidth]{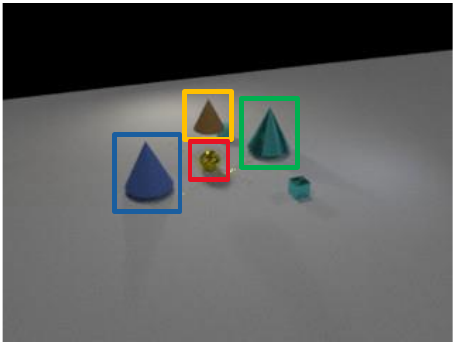} &
\includegraphics[width=0.2\linewidth]{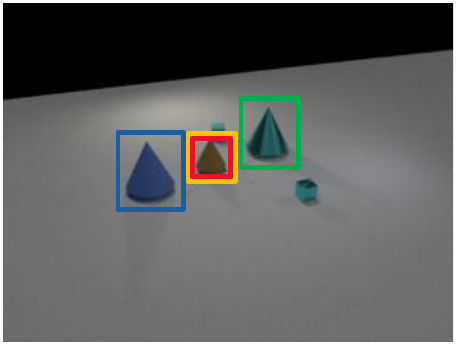} &
\includegraphics[width=0.2\linewidth]{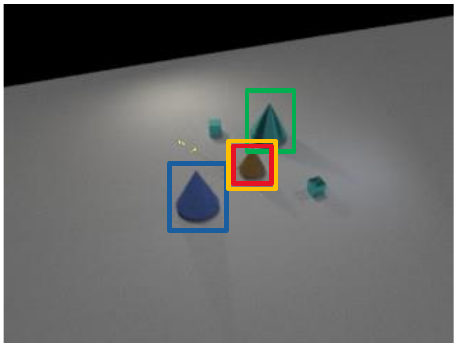}\\

\setulcolor{amber}  \ul{13},{\setulcolor{green} \ul{18}},{\setulcolor{blue} \ul{18}},{\setulcolor{red} \ul{21}} &
\setulcolor{amber}  \ul{21},{\setulcolor{green} \ul{33}},{\setulcolor{blue} \ul{3}},{\setulcolor{red} \ul{21}} &
\setulcolor{amber}  \ul{21},{\setulcolor{green} \ul{33}},{\setulcolor{blue} \ul{3}},{\setulcolor{red} \ul{21}} &
\setulcolor{amber}  \ul{14},{\setulcolor{green} \ul{32}},{\setulcolor{blue} \ul{3}},{\setulcolor{red} \ul{14}}\\
\end{tabularx}
\caption{Example answers to the tracking surrogate task generated by our \model{} model on CATER. Our \model{} model is prompted with the ``<track>'' special token to solve the tracking surrogate task at randomly selected time-steps during training. Object tracks are over the $6 \times 6$ grid on the surface and are highlighted in color.}
\label{tab:cater_qualitative}
\vspace{-0.5 cm}
\end{figure*}

\subsection{CATER}\label{sec:cater}
The CATER (Compositional Actions and TEmporal Reasoning) dataset is designed to test the ability to recognize compositions of object movements that require long-term temporal reasoning. Similar to \cite{abs-2012-08508}, we focus on the hardest task from the CATER dataset, \ie, adversarial target tracking under occlusion and containment.  
This task amounts to predicting the position of a special object, referred to as "snitch”,  at the end of each video sequence. This is challenging as the position of the “snitch” can be occluded or (recursively) contained within other objects at the end of the sequence. This task is posed as a classification problem over a $6 \times 6$ grid. There are two splits of the dataset: static and moving camera -- the latter of which is more challenging as the grid positions can only be predicted by capturing long-term spatiotemporal correlations.

\myparagraph{Surrogate tasks.} We employ (multi-target) tracking as a surrogate task for which the solution can be expressed in the following format: ``<track> \emph{object id$_1$,  object grid position$_1$; $\dots$ ; object id$_n$, object grid position$_n$}'' as shown in \cref{tab:cater_qualitative}. The tracking task includes the medium and large cones in the scene, as these objects can occlude the snitch. We use grid positions instead of bounding boxes (unlike Something-Else) because the final goal is to predict the grid position of the snitch. 
Surrogate tasks are introduced randomly during training with a probability of 30\% after each input frame.

\myparagraph{Baselines and evaluation.} Our \model{} model uses the OPT-125M backbone. We report the results in  \cref{tab:cater_eval}.
To highlight the importance of our surrogate tasks, we consider an ablation without the surrogate tracking task (w/o Surrogate tasks) and to highlight the importance of our Two-stream video encoder we include an ablation with single-stream ViT encoder (w/o Two-stream Encoder).  We train jointly on both the static and moving camera splits, similar to ALOE \cite{abs-2012-08508}. 

The state-of-the-art \blue{powerful transformer-based} IV-CL \citep{abs-2307-08506}, OPNet \citep{abs-2003-10469}, Hopper \citep{ZhouKLNMKG21}, TFC V3D Depthwise \citep{abs-2203-05928} and Loci \citep{abs-2205-13349} report results only on the static camera split. Loci \citep{abs-2205-13349} reports an impressive 90.7\% accuracy on the static camera split, but it is not applicable to the moving camera split due to its static background and camera model.
Our \model{} model outperforms TFC V3D Depthwise \citep{abs-2203-05928} model on the static camera by 4.4\% and ALOE \citep{abs-2012-08508} on the challenging moving camera split by 20.7\%. The large performance gain over the \model{} (w/o Surrogate tasks)  baseline shows the advantage of surrogate tracking tasks, without which the model is not grounded to the motion of the cones and hence fails in cases where the snitch is contained by the cones. Qualitative examples in \cref{tab:cater_qualitative} illustrates that our model is able to successfully track objects in cases of recursive containment and is robust to moving cameras. Finally, the performance advantage over the \model{} (w/o Two-stream encoder) confirms that our \model{} model is able to better capture the motion of the objects. 

\subsection{STAR}
\begin{table*}[!t]
\small
\centering
\caption{Evaluation of our LRR model on STAR (validation set).}
\label{tab:star_eval}
\begin{tabularx}{\linewidth}{@{}X|cccc|c@{}}
\toprule
Method & Int. & Seq. & Pre. & Fea. & Overall $\uparrow$ \\
\midrule
Internvideo \citep{abs-2212-03191} & 62.7 & 65.6 & 54.9 & 51.9 & 58.7 \\
BLIP-2 \citep{0008LSH23} & 65.4 & 69.0 & 59.7 & 54.2 & 62.0 \\
SeViLA \citep{abs-2305-06988} & 63.7 & 70.4 & 63.1 & {62.4} & 64.9  \\
\midrule
\model{} (Ours) & \textbf{73.7} & \textbf{71.0} & \textbf{71.3} & \textbf{65.1} & \textbf{70.5} \\
\midrule
\model{} (w/o Surrogate tasks) &  54.5 & 48.7 & 44.3 & 45.5 & 48.2 \\
\bottomrule
\end{tabularx}
\end{table*}

\blue{STAR \citep{WuYC0G21} is a situated spatio-temporal reasoning benchmark, consisting of questions built upon real-world videos associated with human actions and interactions. The STAR benchmark thus requires an understanding of low-level human motion, actions and object interactions. However, the STAR dataset does not contain dense object annotations, \eg, object tracks, unlike the CATER and Something-Else datasets. It only contains sparse spatio-temporal scene graphs which cover selected keyframes. We introduce object recognition surrogate tasks to localize objects on these keyframes based on the scene graphs. Furthermore, we jointly train our \model{} model to recognize actions from the Kinetics \citep{KayCSZHVVGBNSZ17} and Moments in Time \citep{MonfortVOAZRBYB20} dataset, as well as on surrogate tracking tasks from the Something-Else dataset (see \cref{sec:somethingelse}) to improve the understanding of object interactions. We additionally regularize on text data.}

\blue{Our \model{} model with an OPT-350M LM backbone achieves 70.5\% overall accuracy and significantly outperforms powerful trasnformer-based state of the art models, such as SeViLA \citep{abs-2305-06988}, Internvideo \citep{abs-2212-03191} and BLIP-2 \citep{0008LSH23} as shown in \cref{tab:star_eval}. Following SeViLA \citep{abs-2305-06988}, we report results on the validation set. Note that methods such as SeViLA are trained on a much larger set of image/video - text pairs and, use the 4.1B parameter BLIP as a video encoder and the 3B parameter Flan-T5 XL as an language model (\cf Section 4.4 in \cite{abs-2305-06988}). In contrast, our \model{} model contains fewer parameters and is trained on a much smaller training set. Curicially, our training set includes carefully selected surrogate tasks to endow the model with the requisite low-level visual capabilities. To illustrate this, we include a ablation of our \model{} model without any surrogate tasks (w/o Surrogate tasks). 
}

\blue{In addition to the results on the STAR validation set presented above we also evaluate our LRR model on the STAR challenge leaderboard. The results can be found: \href{https://eval.ai/web/challenges/challenge-page/1325/leaderboard/3328/Mean}{here}, where our LRR model is ranked 1$^\text{st}$ as of January 2024.}

\section{Conclusion}
We show that off-the-shelf LMs can solve complex visual reasoning tasks on videos using our \model{} framework. In this framework, LMs equipped with a two-stream video encoder are supervised and grounded using surrogate tasks. Grounding ensures that the LM can utilize relevant low-level visual cues from in the input video to make predictions.
Grounding predictions to low-level visual cues combined with the high-level reasoning ability of the LM is the key to the success of the model.
Our \model{} model outperforms the state of the art by 6.5\% and 5.3\% on the compositional and systematic splits of the ACRE dataset; by 5.8\% Top-1 accuracy on the compositional split of the Something-Else dataset; by 4.4\% Top-1 accuracy on static camera and 20.7\% Top-1 accuracy on moving camera splits of the CATER dataset; and by 5.6\% overall accuracy on STAR.


\bibliography{iclr2024_conference}

\begin{thebibliography}{81}
\providecommand{\natexlab}[1]{#1}
\providecommand{\url}[1]{\texttt{#1}}
\expandafter\ifx\csname urlstyle\endcsname\relax
  \providecommand{\doi}[1]{doi: #1}\else
  \providecommand{\doi}{doi: \begingroup \urlstyle{rm}\Url}\fi

\bibitem[Alayrac et~al.(2022)Alayrac, Donahue, Luc, Miech, Barr, Hasson, Lenc,
  Mensch, Millican, Reynolds, Ring, Rutherford, Cabi, Han, Gong, Samangooei,
  Monteiro, Menick, Borgeaud, Brock, Nematzadeh, Sharifzadeh, Binkowski,
  Barreira, Vinyals, Zisserman, and Simonyan]{abs-2204-14198}
Jean{-}Baptiste Alayrac, Jeff Donahue, Pauline Luc, Antoine Miech, Iain Barr,
  Yana Hasson, Karel Lenc, Arthur Mensch, Katherine Millican, Malcolm Reynolds,
  Roman Ring, Eliza Rutherford, Serkan Cabi, Tengda Han, Zhitao Gong, Sina
  Samangooei, Marianne Monteiro, Jacob~L. Menick, Sebastian Borgeaud, Andy
  Brock, Aida Nematzadeh, Sahand Sharifzadeh, Mikolaj Binkowski, Ricardo
  Barreira, Oriol Vinyals, Andrew Zisserman, and Kar{\'{e}}n Simonyan.
\newblock Flamingo: a visual language model for few-shot learning.
\newblock In \emph{NeurIPS}, 2022.

\bibitem[Awadalla et~al.(2023)Awadalla, Gao, Gardner, Hessel, Hanafy, Zhu,
  Marathe, Bitton, Gadre, Sagawa, Jitsev, Kornblith, Koh, Ilharco, Wortsman,
  and Schmidt]{abs-2308-01390}
Anas Awadalla, Irena Gao, Josh Gardner, Jack Hessel, Yusuf Hanafy, Wanrong Zhu,
  Kalyani Marathe, Yonatan Bitton, Samir~Yitzhak Gadre, Shiori Sagawa, Jenia
  Jitsev, Simon Kornblith, Pang~Wei Koh, Gabriel Ilharco, Mitchell Wortsman,
  and Ludwig Schmidt.
\newblock Openflamingo: An open-source framework for training large
  autoregressive vision-language models.
\newblock \emph{CoRR}, abs/2308.01390, 2023.

\bibitem[Bai et~al.(2023)Bai, Bai, Yang, Wang, Tan, Wang, Lin, Zhou, and
  Zhou]{abs-2308-12966}
Jinze Bai, Shuai Bai, Shusheng Yang, Shijie Wang, Sinan Tan, Peng Wang, Junyang
  Lin, Chang Zhou, and Jingren Zhou.
\newblock Qwen-vl: {A} frontier large vision-language model with versatile
  abilities.
\newblock \emph{CoRR}, abs/2308.12966, 2023.

\bibitem[Bertasius et~al.(2021)Bertasius, Wang, and Torresani]{BertasiusWT21}
Gedas Bertasius, Heng Wang, and Lorenzo Torresani.
\newblock Is space-time attention all you need for video understanding?
\newblock In Marina Meila and Tong Zhang (eds.), \emph{ICML}, 2021.

\bibitem[Chen et~al.(2018)Chen, Chen, Ma, Jie, and Chua]{ChenCMJC18}
Jingyuan Chen, Xinpeng Chen, Lin Ma, Zequn Jie, and Tat{-}Seng Chua.
\newblock Temporally grounding natural sentence in video.
\newblock In Ellen Riloff, David Chiang, Julia Hockenmaier, and Jun'ichi Tsujii
  (eds.), \emph{ACL}, 2018.

\bibitem[Chen et~al.(2019)Chen, Ma, Chen, Jie, and Luo]{Chen0CJL19}
Jingyuan Chen, Lin Ma, Xinpeng Chen, Zequn Jie, and Jiebo Luo.
\newblock Localizing natural language in videos.
\newblock In \emph{AAAI}, 2019.

\bibitem[Chen et~al.(2022)Chen, Saxena, Li, Lin, Fleet, and
  Hinton]{0007SLLFH22}
Ting Chen, Saurabh Saxena, Lala Li, Tsung{-}Yi Lin, David~J. Fleet, and
  Geoffrey~E. Hinton.
\newblock A unified sequence interface for vision tasks.
\newblock In \emph{NeurIPS}, 2022.

\bibitem[Chiang et~al.(2023)Chiang, Li, Lin, Sheng, Wu, Zhang, Zheng, Zhuang,
  Zhuang, Gonzalez, Stoica, and Xing]{vicuna2023}
Wei-Lin Chiang, Zhuohan Li, Zi~Lin, Ying Sheng, Zhanghao Wu, Hao Zhang, Lianmin
  Zheng, Siyuan Zhuang, Yonghao Zhuang, Joseph~E. Gonzalez, Ion Stoica, and
  Eric~P. Xing.
\newblock Vicuna: An open-source chatbot impressing gpt-4 with 90\%* chatgpt
  quality, March 2023.
\newblock URL \url{https://lmsys.org/blog/2023-03-30-vicuna/}.

\bibitem[Cobbe et~al.(2021)Cobbe, Kosaraju, Bavarian, Hilton, Nakano, Hesse,
  and Schulman]{abs-2110-14168}
Karl Cobbe, Vineet Kosaraju, Mohammad Bavarian, Jacob Hilton, Reiichiro Nakano,
  Christopher Hesse, and John Schulman.
\newblock Training verifiers to solve math word problems.
\newblock \emph{CoRR}, abs/2110.14168, 2021.

\bibitem[Cubuk et~al.(2020)Cubuk, Zoph, Shlens, and Le]{CubukZSL20}
Ekin~D. Cubuk, Barret Zoph, Jonathon Shlens, and Quoc~V. Le.
\newblock Randaugment: Practical automated data augmentation with a reduced
  search space.
\newblock In \emph{CVPR Workshops}, 2020.

\bibitem[Dai et~al.(2023)Dai, Li, Li, Tiong, Zhao, Wang, Li, Fung, and
  Hoi]{abs-2305-06500}
Wenliang Dai, Junnan Li, Dongxu Li, Anthony Meng~Huat Tiong, Junqi Zhao,
  Weisheng Wang, Boyang Li, Pascale Fung, and Steven C.~H. Hoi.
\newblock Instructblip: Towards general-purpose vision-language models with
  instruction tuning.
\newblock In \emph{NeurIPS}, 2023.

\bibitem[Deng et~al.(2021)Deng, Yang, Chen, Zhou, and Li]{DengYCZL21}
Jiajun Deng, Zhengyuan Yang, Tianlang Chen, Wengang Zhou, and Houqiang Li.
\newblock Transvg: End-to-end visual grounding with transformers.
\newblock In \emph{ICCV}, 2021.

\bibitem[Ding et~al.(2021)Ding, Hill, Santoro, Reynolds, and
  Botvinick]{abs-2012-08508}
David Ding, Felix Hill, Adam Santoro, Malcolm Reynolds, and Matt~M. Botvinick.
\newblock Attention over learned object embeddings enables complex visual
  reasoning.
\newblock In \emph{NeurIPS}, 2021.

\bibitem[Dosovitskiy et~al.(2021)Dosovitskiy, Beyer, Kolesnikov, Weissenborn,
  Zhai, Unterthiner, Dehghani, Minderer, Heigold, Gelly, Uszkoreit, and
  Houlsby]{DosovitskiyB0WZ21}
Alexey Dosovitskiy, Lucas Beyer, Alexander Kolesnikov, Dirk Weissenborn,
  Xiaohua Zhai, Thomas Unterthiner, Mostafa Dehghani, Matthias Minderer, Georg
  Heigold, Sylvain Gelly, Jakob Uszkoreit, and Neil Houlsby.
\newblock An image is worth 16x16 words: Transformers for image recognition at
  scale.
\newblock In \emph{ICLR}, 2021.

\bibitem[Driess et~al.(2023)Driess, Xia, Sajjadi, Lynch, Chowdhery, Ichter,
  Wahid, Tompson, Vuong, Yu, Huang, Chebotar, Sermanet, Duckworth, Levine,
  Vanhoucke, Hausman, Toussaint, Greff, Zeng, Mordatch, and
  Florence]{abs-2303-03378}
Danny Driess, Fei Xia, Mehdi S.~M. Sajjadi, Corey Lynch, Aakanksha Chowdhery,
  Brian Ichter, Ayzaan Wahid, Jonathan Tompson, Quan Vuong, Tianhe Yu, Wenlong
  Huang, Yevgen Chebotar, Pierre Sermanet, Daniel Duckworth, Sergey Levine,
  Vincent Vanhoucke, Karol Hausman, Marc Toussaint, Klaus Greff, Andy Zeng,
  Igor Mordatch, and Pete Florence.
\newblock Palm-e: An embodied multimodal language model.
\newblock \emph{CoRR}, abs/2303.03378, 2023.

\bibitem[Gao et~al.(2021)Gao, Biderman, Black, Golding, Hoppe, Foster, Phang,
  He, Thite, Nabeshima, Presser, and Leahy]{abs-2101-00027}
Leo Gao, Stella Biderman, Sid Black, Laurence Golding, Travis Hoppe, Charles
  Foster, Jason Phang, Horace He, Anish Thite, Noa Nabeshima, Shawn Presser,
  and Connor Leahy.
\newblock The pile: An 800gb dataset of diverse text for language modeling.
\newblock \emph{CoRR}, abs/2101.00027, 2021.

\bibitem[Gendron et~al.(2023)Gendron, Bao, Witbrock, and
  Dobbie]{abs-2305-19555}
Ga{\"{e}}l Gendron, Qiming Bao, Michael Witbrock, and Gillian Dobbie.
\newblock Large language models are not abstract reasoners.
\newblock \emph{CoRR}, abs/2305.19555, 2023.

\bibitem[Girdhar \& Ramanan(2020)Girdhar and Ramanan]{GirdharR20}
Rohit Girdhar and Deva Ramanan.
\newblock {CATER:} {A} diagnostic dataset for compositional actions {\&}
  temporal reasoning.
\newblock In \emph{ICLR}, 2020.

\bibitem[Goyal et~al.(2017)Goyal, Kahou, Michalski, Materzynska, Westphal, Kim,
  Haenel, Fr{\"{u}}nd, Yianilos, Mueller{-}Freitag, Hoppe, Thurau, Bax, and
  Memisevic]{GoyalKMMWKHFYMH17}
Raghav Goyal, Samira~Ebrahimi Kahou, Vincent Michalski, Joanna Materzynska,
  Susanne Westphal, Heuna Kim, Valentin Haenel, Ingo Fr{\"{u}}nd, Peter
  Yianilos, Moritz Mueller{-}Freitag, Florian Hoppe, Christian Thurau, Ingo
  Bax, and Roland Memisevic.
\newblock The "something something" video database for learning and evaluating
  visual common sense.
\newblock In \emph{ICCV}, 2017.

\bibitem[Gupta \& Kembhavi(2022)Gupta and Kembhavi]{abs-2211-11559}
Tanmay Gupta and Aniruddha Kembhavi.
\newblock Visual programming: Compositional visual reasoning without training.
\newblock \emph{CoRR}, abs/2211.11559, 2022.

\bibitem[He et~al.(2019)He, Zhao, Huang, Li, Liu, and Wen]{HeZHLLW19}
Dongliang He, Xiang Zhao, Jizhou Huang, Fu~Li, Xiao Liu, and Shilei Wen.
\newblock Read, watch, and move: Reinforcement learning for temporally
  grounding natural language descriptions in videos.
\newblock In \emph{AAAI}, 2019.

\bibitem[Hu et~al.(2017)Hu, Andreas, Rohrbach, Darrell, and Saenko]{HuARDS17}
Ronghang Hu, Jacob Andreas, Marcus Rohrbach, Trevor Darrell, and Kate Saenko.
\newblock Learning to reason: End-to-end module networks for visual question
  answering.
\newblock In \emph{ICCV}, 2017.

\bibitem[Huang et~al.(2018)Huang, Buch, Dery, Garg, Fei{-}Fei, and
  Niebles]{HuangBDGFN18}
De{-}An Huang, Shyamal Buch, Lucio~M. Dery, Animesh Garg, Li~Fei{-}Fei, and
  Juan~Carlos Niebles.
\newblock Finding "it": Weakly-supervised reference-aware visual grounding in
  instructional videos.
\newblock In \emph{CVPR}, 2018.

\bibitem[Hudson \& Manning(2018)Hudson and Manning]{HudsonM18}
Drew~A. Hudson and Christopher~D. Manning.
\newblock Compositional attention networks for machine reasoning.
\newblock In \emph{ICLR}, 2018.

\bibitem[Jang et~al.(2023)Jang, Park, Kim, Kwon, and Sohn]{abs-2308-06947}
Jinhyun Jang, Jungin Park, Jin Kim, Hyeongjun Kwon, and Kwanghoon Sohn.
\newblock Knowing where to focus: Event-aware transformer for video grounding.
\newblock \emph{CoRR}, abs/2308.06947, 2023.

\bibitem[Jin et~al.(2022)Jin, Li, Yuan, and Mu]{JinLYM22}
Yang Jin, Yongzhi Li, Zehuan Yuan, and Yadong Mu.
\newblock Embracing consistency: {A} one-stage approach for spatio-temporal
  video grounding.
\newblock In \emph{NeurIPS}, 2022.

\bibitem[Kamath et~al.(2021)Kamath, Singh, LeCun, Synnaeve, Misra, and
  Carion]{KamathSLSMC21}
Aishwarya Kamath, Mannat Singh, Yann LeCun, Gabriel Synnaeve, Ishan Misra, and
  Nicolas Carion.
\newblock {MDETR} - modulated detection for end-to-end multi-modal
  understanding.
\newblock In \emph{ICCV}, 2021.

\bibitem[Kay et~al.(2017)Kay, Carreira, Simonyan, Zhang, Hillier,
  Vijayanarasimhan, Viola, Green, Back, Natsev, Suleyman, and
  Zisserman]{KayCSZHVVGBNSZ17}
Will Kay, Jo{\~{a}}o Carreira, Karen Simonyan, Brian Zhang, Chloe Hillier,
  Sudheendra Vijayanarasimhan, Fabio Viola, Tim Green, Trevor Back, Paul
  Natsev, Mustafa Suleyman, and Andrew Zisserman.
\newblock The kinetics human action video dataset.
\newblock \emph{CoRR}, abs/1705.06950, 2017.

\bibitem[Koh et~al.(2023)Koh, Salakhutdinov, and Fried]{abs-2301-13823}
Jing~Yu Koh, Ruslan Salakhutdinov, and Daniel Fried.
\newblock Grounding language models to images for multimodal generation.
\newblock \emph{CoRR}, abs/2301.13823, 2023.

\bibitem[Li et~al.(2023{\natexlab{a}})Li, Li, Savarese, and Hoi]{0008LSH23}
Junnan Li, Dongxu Li, Silvio Savarese, and Steven C.~H. Hoi.
\newblock {BLIP-2:} bootstrapping language-image pre-training with frozen image
  encoders and large language models.
\newblock In \emph{ICML}, 2023{\natexlab{a}}.

\bibitem[Li et~al.(2023{\natexlab{b}})Li, He, Wang, Li, Wang, Luo, Wang, Wang,
  and Qiao]{abs-2305-06355}
Kunchang Li, Yinan He, Yi~Wang, Yizhuo Li, Wenhai Wang, Ping Luo, Yali Wang,
  Limin Wang, and Yu~Qiao.
\newblock Videochat: Chat-centric video understanding.
\newblock \emph{CoRR}, abs/2305.06355, 2023{\natexlab{b}}.

\bibitem[Li et~al.(2023{\natexlab{c}})Li, Wang, Zhang, Miao, Zhao, Zhang, Ji,
  and Wu]{LiWZMZZJW23}
Mengze Li, Han Wang, Wenqiao Zhang, Jiaxu Miao, Zhou Zhao, Shengyu Zhang, Wei
  Ji, and Fei Wu.
\newblock {WINNER:} weakly-supervised hierarchical decomposition and alignment
  for spatio-temporal video grounding.
\newblock In \emph{CVPR}, 2023{\natexlab{c}}.

\bibitem[Li et~al.(2021)Li, Stengel{-}Eskin, Zhang, Xie, Tran, Durme, and
  Yuille]{LiSZXTDY21}
Zhuowan Li, Elias Stengel{-}Eskin, Yixiao Zhang, Cihang Xie, Quan Tran,
  Benjamin~Van Durme, and Alan~L. Yuille.
\newblock Calibrating concepts and operations: Towards symbolic reasoning on
  real images.
\newblock In \emph{ICCV}, 2021.

\bibitem[Loshchilov \& Hutter(2019)Loshchilov and Hutter]{LoshchilovH19}
Ilya Loshchilov and Frank Hutter.
\newblock Decoupled weight decay regularization.
\newblock In \emph{ICLR}, 2019.

\bibitem[Lu et~al.(2023)Lu, Peng, Cheng, Galley, Chang, Wu, Zhu, and
  Gao]{abs-2304-09842}
Pan Lu, Baolin Peng, Hao Cheng, Michel Galley, Kai{-}Wei Chang, Ying~Nian Wu,
  Song{-}Chun Zhu, and Jianfeng Gao.
\newblock Chameleon: Plug-and-play compositional reasoning with large language
  models.
\newblock \emph{CoRR}, abs/2304.09842, 2023.

\bibitem[Luo et~al.(2020)Luo, Zhou, Sun, Cao, Wu, Deng, and Ji]{LuoZSCWDJ20}
Gen Luo, Yiyi Zhou, Xiaoshuai Sun, Liujuan Cao, Chenglin Wu, Cheng Deng, and
  Rongrong Ji.
\newblock Multi-task collaborative network for joint referring expression
  comprehension and segmentation.
\newblock In \emph{CVPR}, 2020.

\bibitem[Luo et~al.(2023)Luo, Zhao, Yang, Dong, Qiu, Lu, Wang, and
  Wei]{abs-2306-07207}
Ruipu Luo, Ziwang Zhao, Min Yang, Junwei Dong, Minghui Qiu, Pengcheng Lu, Tao
  Wang, and Zhongyu Wei.
\newblock Valley: Video assistant with large language model enhanced ability.
\newblock \emph{CoRR}, abs/2306.07207, 2023.

\bibitem[Maaz et~al.(2023)Maaz, Rasheed, Khan, and Khan]{abs-2306-05424}
Muhammad Maaz, Hanoona~Abdul Rasheed, Salman~H. Khan, and Fahad~Shahbaz Khan.
\newblock Video-chatgpt: Towards detailed video understanding via large vision
  and language models.
\newblock \emph{CoRR}, abs/2306.05424, 2023.

\bibitem[Mahajan \& Roth(2020)Mahajan and Roth]{Mahajan020}
Shweta Mahajan and Stefan Roth.
\newblock Diverse image captioning with context-object split latent spaces.
\newblock In \emph{NeurIPS}, 2020.

\bibitem[Materzynska et~al.(2020)Materzynska, Xiao, Herzig, Xu, Wang, and
  Darrell]{MaterzynskaXHXW20}
Joanna Materzynska, Tete Xiao, Roei Herzig, Huijuan Xu, Xiaolong Wang, and
  Trevor Darrell.
\newblock Something-else: Compositional action recognition with
  spatial-temporal interaction networks.
\newblock In \emph{CVPR}, 2020.

\bibitem[Mondal et~al.(2023)Mondal, Webb, and Cohen]{abs-2303-02260}
Shanka~Subhra Mondal, Taylor Webb, and Jonathan~D. Cohen.
\newblock Learning to reason over visual objects.
\newblock \emph{CoRR}, abs/2303.02260, 2023.

\bibitem[Monfort et~al.(2020)Monfort, Vondrick, Oliva, Andonian, Zhou,
  Ramakrishnan, Bargal, Yan, Brown, Fan, and Gutfreund]{MonfortVOAZRBYB20}
Mathew Monfort, Carl Vondrick, Aude Oliva, Alex Andonian, Bolei Zhou, Kandan
  Ramakrishnan, Sarah~Adel Bargal, Tom Yan, Lisa~M. Brown, Quanfu Fan, and Dan
  Gutfreund.
\newblock Moments in time dataset: One million videos for event understanding.
\newblock \emph{{IEEE} Trans. Pattern Anal. Mach. Intell.}, 42\penalty0
  (2):\penalty0 502--508, 2020.

\bibitem[OpenAI(2023)]{abs-2303-08774}
OpenAI.
\newblock {GPT-4} technical report.
\newblock \emph{CoRR}, abs/2303.08774, 2023.

\bibitem[Peng et~al.(2023)Peng, Wang, Dong, Hao, Huang, Ma, and
  Wei]{abs-2306-14824}
Zhiliang Peng, Wenhui Wang, Li~Dong, Yaru Hao, Shaohan Huang, Shuming Ma, and
  Furu Wei.
\newblock Kosmos-2: Grounding multimodal large language models to the world.
\newblock \emph{CoRR}, abs/2306.14824, 2023.

\bibitem[Radford et~al.(2021)Radford, Kim, Hallacy, Ramesh, Goh, Agarwal,
  Sastry, Askell, Mishkin, Clark, Krueger, and Sutskever]{RadfordKHRGASAM21}
Alec Radford, Jong~Wook Kim, Chris Hallacy, Aditya Ramesh, Gabriel Goh,
  Sandhini Agarwal, Girish Sastry, Amanda Askell, Pamela Mishkin, Jack Clark,
  Gretchen Krueger, and Ilya Sutskever.
\newblock Learning transferable visual models from natural language
  supervision.
\newblock In \emph{ICML}, 2021.

\bibitem[Rahaman et~al.(2021)Rahaman, Gondal, Joshi, Gehler, Bengio, Locatello,
  and Sch{\"{o}}lkopf]{RahamanGJGBLS21}
Nasim Rahaman, Muhammad~Waleed Gondal, Shruti Joshi, Peter~V. Gehler, Yoshua
  Bengio, Francesco Locatello, and Bernhard Sch{\"{o}}lkopf.
\newblock Dynamic inference with neural interpreters.
\newblock In \emph{NeurIPS}, 2021.

\bibitem[Ren et~al.(2017)Ren, He, Girshick, and Sun]{RenHG017}
Shaoqing Ren, Kaiming He, Ross~B. Girshick, and Jian Sun.
\newblock Faster {R-CNN:} towards real-time object detection with region
  proposal networks.
\newblock \emph{{IEEE} Trans. Pattern Anal. Mach. Intell.}, 39\penalty0
  (6):\penalty0 1137--1149, 2017.

\bibitem[Santoro et~al.(2017)Santoro, Raposo, Barrett, Malinowski, Pascanu,
  Battaglia, and Lillicrap]{SantoroRBMPBL17}
Adam Santoro, David Raposo, David G.~T. Barrett, Mateusz Malinowski, Razvan
  Pascanu, Peter~W. Battaglia, and Tim Lillicrap.
\newblock A simple neural network module for relational reasoning.
\newblock In \emph{NeurIPS}, 2017.

\bibitem[Scao et~al.(2022)Scao, Fan, Akiki, Pavlick, Ilic, Hesslow,
  Castagn{\'{e}}, Luccioni, Yvon, Gall{\'{e}}, Tow, Rush, Biderman, Webson,
  Ammanamanchi, Wang, Sagot, Muennighoff, del Moral, Ruwase, Bawden, Bekman,
  McMillan{-}Major, Beltagy, Nguyen, Saulnier, Tan, Suarez, Sanh,
  Lauren{\c{c}}on, Jernite, Launay, Mitchell, Raffel, Gokaslan, Simhi, Soroa,
  Aji, Alfassy, Rogers, Nitzav, Xu, Mou, Emezue, Klamm, Leong, van Strien,
  Adelani, and et~al.]{abs-2211-05100}
Teven~Le Scao, Angela Fan, Christopher Akiki, Ellie Pavlick, Suzana Ilic,
  Daniel Hesslow, Roman Castagn{\'{e}}, Alexandra~Sasha Luccioni,
  Fran{\c{c}}ois Yvon, Matthias Gall{\'{e}}, Jonathan Tow, Alexander~M. Rush,
  Stella Biderman, Albert Webson, Pawan~Sasanka Ammanamanchi, Thomas Wang,
  Beno{\^{\i}}t Sagot, Niklas Muennighoff, Albert~Villanova del Moral, Olatunji
  Ruwase, Rachel Bawden, Stas Bekman, Angelina McMillan{-}Major, Iz~Beltagy,
  Huu Nguyen, Lucile Saulnier, Samson Tan, Pedro~Ortiz Suarez, Victor Sanh,
  Hugo Lauren{\c{c}}on, Yacine Jernite, Julien Launay, Margaret Mitchell, Colin
  Raffel, Aaron Gokaslan, Adi Simhi, Aitor Soroa, Alham~Fikri Aji, Amit
  Alfassy, Anna Rogers, Ariel~Kreisberg Nitzav, Canwen Xu, Chenghao Mou, Chris
  Emezue, Christopher Klamm, Colin Leong, Daniel van Strien, David~Ifeoluwa
  Adelani, and et~al.
\newblock {BLOOM:} {A} 176b-parameter open-access multilingual language model.
\newblock \emph{CoRR}, abs/2211.05100, 2022.

\bibitem[Shamsian et~al.(2020)Shamsian, Kleinfeld, Globerson, and
  Chechik]{abs-2003-10469}
Aviv Shamsian, Ofri Kleinfeld, Amir Globerson, and Gal Chechik.
\newblock Learning object permanence from video.
\newblock In \emph{ECCV}, 2020.

\bibitem[Shi et~al.(2019)Shi, Xu, Gong, and Xu]{0005XGX19}
Jing Shi, Jia Xu, Boqing Gong, and Chenliang Xu.
\newblock Not all frames are equal: Weakly-supervised video grounding wit
  contextual similarity and visual clustering losses.
\newblock In \emph{CVPR}, 2019.

\bibitem[Simonyan \& Zisserman(2014)Simonyan and Zisserman]{SimonyanZ14}
Karen Simonyan and Andrew Zisserman.
\newblock Two-stream convolutional networks for action recognition in videos.
\newblock In Zoubin Ghahramani, Max Welling, Corinna Cortes, Neil~D. Lawrence,
  and Kilian~Q. Weinberger (eds.), \emph{NeurIPS}, 2014.

\bibitem[Strub et~al.(2018)Strub, Seurin, Perez, de~Vries, Mary, Preux,
  Courville, and Pietquin]{StrubSPVMPCP18}
Florian Strub, Mathieu Seurin, Ethan Perez, Harm de~Vries, J{\'{e}}r{\'{e}}mie
  Mary, Philippe Preux, Aaron~C. Courville, and Olivier Pietquin.
\newblock Visual reasoning with multi-hop feature modulation.
\newblock In \emph{ECCV}, 2018.

\bibitem[Su et~al.(2021)Su, Yu, and Xu]{SuY021}
Rui Su, Qian Yu, and Dong Xu.
\newblock Stvgbert: {A} visual-linguistic transformer based framework for
  spatio-temporal video grounding.
\newblock In \emph{ICCV}, 2021.

\bibitem[Sun et~al.(2023)Sun, Luo, Zhou, Arnab, and Schmid]{abs-2307-08506}
Chen Sun, Calvin Luo, Xingyi Zhou, Anurag Arnab, and Cordelia Schmid.
\newblock Does visual pretraining help end-to-end reasoning?
\newblock \emph{CoRR}, abs/2307.08506, 2023.

\bibitem[Sur{\'{\i}}s et~al.(2023)Sur{\'{\i}}s, Menon, and
  Vondrick]{abs-2303-08128}
D{\'{\i}}dac Sur{\'{\i}}s, Sachit Menon, and Carl Vondrick.
\newblock Vipergpt: Visual inference via python execution for reasoning.
\newblock \emph{CoRR}, abs/2303.08128, 2023.

\bibitem[Tang et~al.(2017)Tang, Andriluka, Andres, and Schiele]{TangAAS17}
Siyu Tang, Mykhaylo Andriluka, Bjoern Andres, and Bernt Schiele.
\newblock Multiple people tracking by lifted multicut and person
  re-identification.
\newblock In \emph{CVPR}, 2017.

\bibitem[Touvron et~al.(2023)Touvron, Lavril, Izacard, Martinet, Lachaux,
  Lacroix, Rozi{\`{e}}re, Goyal, Hambro, Azhar, Rodriguez, Joulin, Grave, and
  Lample]{abs-2302-13971}
Hugo Touvron, Thibaut Lavril, Gautier Izacard, Xavier Martinet, Marie{-}Anne
  Lachaux, Timoth{\'{e}}e Lacroix, Baptiste Rozi{\`{e}}re, Naman Goyal, Eric
  Hambro, Faisal Azhar, Aur{\'{e}}lien Rodriguez, Armand Joulin, Edouard Grave,
  and Guillaume Lample.
\newblock Llama: Open and efficient foundation language models.
\newblock \emph{CoRR}, abs/2302.13971, 2023.

\bibitem[Traub et~al.(2023)Traub, Otte, Menge, Karlbauer, Th{\"{u}}mmel, and
  Butz]{abs-2205-13349}
Manuel Traub, Sebastian Otte, Tobias Menge, Matthias Karlbauer, Jannik
  Th{\"{u}}mmel, and Martin~V. Butz.
\newblock Learning what and where - unsupervised disentangling location and
  identity tracking.
\newblock In \emph{ICLR}, 2023.

\bibitem[Vasudevan et~al.(2018)Vasudevan, Dai, and Gool]{VasudevanDG18}
Arun~Balajee Vasudevan, Dengxin Dai, and Luc~Van Gool.
\newblock Object referring in videos with language and human gaze.
\newblock In \emph{CVPR}, 2018.

\bibitem[Vaswani et~al.(2017)Vaswani, Shazeer, Parmar, Uszkoreit, Jones, Gomez,
  Kaiser, and Polosukhin]{VaswaniSPUJGKP17}
Ashish Vaswani, Noam Shazeer, Niki Parmar, Jakob Uszkoreit, Llion Jones,
  Aidan~N. Gomez, Lukasz Kaiser, and Illia Polosukhin.
\newblock Attention is all you need.
\newblock In \emph{NeurIPS}, 2017.

\bibitem[Wang et~al.(2022)Wang, Li, Li, He, Huang, Zhao, Zhang, Xu, Liu, Wang,
  Xing, Chen, Pan, Yu, Wang, Wang, and Qiao]{abs-2212-03191}
Yi~Wang, Kunchang Li, Yizhuo Li, Yinan He, Bingkun Huang, Zhiyu Zhao, Hongjie
  Zhang, Jilan Xu, Yi~Liu, Zun Wang, Sen Xing, Guo Chen, Junting Pan, Jiashuo
  Yu, Yali Wang, Limin Wang, and Yu~Qiao.
\newblock Internvideo: General video foundation models via generative and
  discriminative learning.
\newblock \emph{CoRR}, abs/2212.03191, 2022.

\bibitem[Wei et~al.(2022)Wei, Wang, Schuurmans, Bosma, brian ichter, Xia, Chi,
  Le, and Zhou]{wei2022chain}
Jason Wei, Xuezhi Wang, Dale Schuurmans, Maarten Bosma, brian ichter, Fei Xia,
  Ed~H. Chi, Quoc~V Le, and Denny Zhou.
\newblock Chain of thought prompting elicits reasoning in large language
  models.
\newblock In \emph{NeurIPS}, 2022.

\bibitem[Wu et~al.(2021)Wu, Yu, Chen, Tenenbaum, and Gan]{WuYC0G21}
Bo~Wu, Shoubin Yu, Zhenfang Chen, Josh Tenenbaum, and Chuang Gan.
\newblock {STAR:} {A} benchmark for situated reasoning in real-world videos.
\newblock In Joaquin Vanschoren and Sai{-}Kit Yeung (eds.), \emph{NeurIPS},
  2021.

\bibitem[Xiao et~al.(2023)Xiao, Yao, Li, and Chua]{abs-2309-01327}
Junbin Xiao, Angela Yao, Yicong Li, and Tat{-}Seng Chua.
\newblock Can {I} trust your answer? visually grounded video question
  answering.
\newblock \emph{CoRR}, abs/2309.01327, 2023.

\bibitem[Yang et~al.(2022)Yang, Miech, Sivic, Laptev, and Schmid]{YangMSLS22}
Antoine Yang, Antoine Miech, Josef Sivic, Ivan Laptev, and Cordelia Schmid.
\newblock Tubedetr: Spatio-temporal video grounding with transformers.
\newblock In \emph{CVPR}, 2022.

\bibitem[Yang et~al.(2019)Yang, Gong, Wang, Huang, Yu, and Luo]{YangGWHYL19}
Zhengyuan Yang, Boqing Gong, Liwei Wang, Wenbing Huang, Dong Yu, and Jiebo Luo.
\newblock A fast and accurate one-stage approach to visual grounding.
\newblock In \emph{ICCV}, 2019.

\bibitem[Yang et~al.(2020)Yang, Chen, Wang, and Luo]{YangC0L20}
Zhengyuan Yang, Tianlang Chen, Liwei Wang, and Jiebo Luo.
\newblock Improving one-stage visual grounding by recursive sub-query
  construction.
\newblock In Andrea Vedaldi, Horst Bischof, Thomas Brox, and Jan{-}Michael
  Frahm (eds.), \emph{ECCV}, 2020.

\bibitem[Yao et~al.(2018)Yao, Xu, Wang, and Xu]{YaoXWX18}
Yiqun Yao, Jiaming Xu, Feng Wang, and Bo~Xu.
\newblock Cascaded mutual modulation for visual reasoning.
\newblock In \emph{EMNLP}, 2018.

\bibitem[Ye et~al.(2022)Ye, Shen, Lin, Xiang, Shao, and Hoi]{YeSLXSH22}
Mang Ye, Jianbing Shen, Gaojie Lin, Tao Xiang, Ling Shao, and Steven C.~H. Hoi.
\newblock Deep learning for person re-identification: {A} survey and outlook.
\newblock \emph{{IEEE} Trans. Pattern Anal. Mach. Intell.}, 44\penalty0
  (6):\penalty0 2872--2893, 2022.

\bibitem[Yu et~al.(2023)Yu, Cho, Yadav, and Bansal]{abs-2305-06988}
Shoubin Yu, Jaemin Cho, Prateek Yadav, and Mohit Bansal.
\newblock Self-chained image-language model for video localization and question
  answering.
\newblock \emph{CoRR}, abs/2305.06988, 2023.

\bibitem[Zeng et~al.(2020)Zeng, Xu, Huang, Chen, Tan, and Gan]{ZengXHCTG20}
Runhao Zeng, Haoming Xu, Wenbing Huang, Peihao Chen, Mingkui Tan, and Chuang
  Gan.
\newblock Dense regression network for video grounding.
\newblock In \emph{CVPR}, 2020.

\bibitem[Zhang et~al.(2021)Zhang, Jia, Edmonds, Zhu, and Zhu]{0017JEZZ21}
Chi Zhang, Baoxiong Jia, Mark Edmonds, Song{-}Chun Zhu, and Yixin Zhu.
\newblock {ACRE:} abstract causal reasoning beyond covariation.
\newblock In \emph{CVPR}, 2021.

\bibitem[Zhang et~al.(2022{\natexlab{a}})Zhang, Xie, Jia, Wu, Zhu, and
  Zhu]{ZhangXJWZZ22}
Chi Zhang, Sirui Xie, Baoxiong Jia, Ying~Nian Wu, Song{-}Chun Zhu, and Yixin
  Zhu.
\newblock Learning algebraic representation for systematic generalization in
  abstract reasoning.
\newblock In \emph{ECCV}, 2022{\natexlab{a}}.

\bibitem[Zhang et~al.(2019)Zhang, Dai, Wang, Wang, and Davis]{ZhangDWWD19}
Da~Zhang, Xiyang Dai, Xin Wang, Yuan{-}Fang Wang, and Larry~S. Davis.
\newblock {MAN:} moment alignment network for natural language moment retrieval
  via iterative graph adjustment.
\newblock In \emph{CVPR}, 2019.

\bibitem[Zhang et~al.(2020)Zhang, Sun, Jing, and Zhou]{ZhangSJZ20}
Hao Zhang, Aixin Sun, Wei Jing, and Joey~Tianyi Zhou.
\newblock Span-based localizing network for natural language video
  localization.
\newblock In Dan Jurafsky, Joyce Chai, Natalie Schluter, and Joel~R. Tetreault
  (eds.), \emph{ACL}, 2020.

\bibitem[Zhang et~al.(2023{\natexlab{a}})Zhang, Han, Zhou, Hu, Yan, Lu, Li,
  Gao, and Qiao]{zhang2023llamaadapter}
Renrui Zhang, Jiaming Han, Aojun Zhou, Xiangfei Hu, Shilin Yan, Pan Lu,
  Hongsheng Li, Peng Gao, and Yu~Qiao.
\newblock Llama-adapter: Efficient fine-tuning of language models with
  zero-init attention.
\newblock \emph{CoRR}, abs/2303.16199, 2023{\natexlab{a}}.

\bibitem[Zhang(2022)]{abs-2203-05928}
Shiwen Zhang.
\newblock Tfcnet: Temporal fully connected networks for static unbiased
  temporal reasoning.
\newblock \emph{CoRR}, 2022.

\bibitem[Zhang et~al.(2022{\natexlab{b}})Zhang, Roller, Goyal, Artetxe, Chen,
  Chen, Dewan, Diab, Li, Lin, Mihaylov, Ott, Shleifer, Shuster, Simig, Koura,
  Sridhar, Wang, and Zettlemoyer]{abs-2205-01068}
Susan Zhang, Stephen Roller, Naman Goyal, Mikel Artetxe, Moya Chen, Shuohui
  Chen, Christopher Dewan, Mona~T. Diab, Xian Li, Xi~Victoria Lin, Todor
  Mihaylov, Myle Ott, Sam Shleifer, Kurt Shuster, Daniel Simig, Punit~Singh
  Koura, Anjali Sridhar, Tianlu Wang, and Luke Zettlemoyer.
\newblock {OPT:} open pre-trained transformer language models.
\newblock \emph{CoRR}, abs/2205.01068, 2022{\natexlab{b}}.

\bibitem[Zhang et~al.(2023{\natexlab{b}})Zhang, Zhang, Li, Zhao, Karypis, and
  Smola]{abs-2302-00923}
Zhuosheng Zhang, Aston Zhang, Mu~Li, Hai Zhao, George Karypis, and Alex Smola.
\newblock Multimodal chain-of-thought reasoning in language models.
\newblock \emph{CoRR}, abs/2302.00923, 2023{\natexlab{b}}.

\bibitem[Zhou et~al.(2021)Zhou, Kadav, Lai, Niculescu{-}Mizil, Min, Kapadia,
  and Graf]{ZhouKLNMKG21}
Honglu Zhou, Asim Kadav, Farley Lai, Alexandru Niculescu{-}Mizil,
  Martin~Renqiang Min, Mubbasir Kapadia, and Hans~Peter Graf.
\newblock Hopper: Multi-hop transformer for spatiotemporal reasoning.
\newblock In \emph{ICLR}, 2021.

\end{thebibliography}
\bibliographystyle{iclr2024_conference}
\clearpage
\appendix
\section*{\centering Appendix}
\section{Overview}
Here we provide, 
\begin{enumerate*}
    \item Results on a ``generalist'' \model{} model.
    \item Surrogate tasks and overfitting.
    \item Additional training details including hyper-parameters used for evaluation in Section 4 of the main paper across ACRE, Something-Else and CATER datasets.
    \item Additional experiments to compare our \model{} approach to ``Chain-of-thought'' \citep{wei2022chain} based approaches.
    \item Additional qualitative examples across ACRE, Something-Else and CATER datasets.
\end{enumerate*}

\subsection{Generalist Model}\label{sec:joint}
The results in the main paper show that we can leverage the reasoning capabilities of pre-trained LMs to create ``specialist'' models with state of the art performance on a variety of visual reasoning tasks using end-to-end learning. 
Next, as both the input modality and the acquired visual skills are generic and in principle applicable across tasks, we train the 
model with OPT-125M backbone 
jointly on the two video-based reasoning tasks discussed above, CATER and Something-Else.\footnote{\scriptsize{It may be possible to include datasets like ACRE by treating images as still videos, but we leave this for future work.}} 
This is a step towards developing a grounded ``generalist'' for 
performing reasoning in videos. 
Despite commonalities in underlying low-level features, the diverse nature of reasoning objectives in the two tasks makes this highly challenging. 
As shown in \Cref{tab:joint_training}, our \model{} (Joint) model outperforms even the dataset specific fine-tuned state of the art models (SOTA) \citep{MaterzynskaXHXW20,abs-2203-05928,abs-2012-08508} on these highly diverse visual reasoning tasks.

\begin{table*}[h]
\centering
\scriptsize
\caption{Evaluation of our \model{} model trained jointly on CATER and Something-Else.}
\vspace{-0.15cm}
\label{tab:joint_training}
\begin{tabularx}{\linewidth}{@{}Xcccccc@{}}
\toprule
& \multicolumn{2}{c}{Something-Else} & \multicolumn{2}{c}{CATER: Static Camera} & \multicolumn{2}{c}{CATER: Moving Camera} \\
\cmidrule(l{0.7em}){2-3} \cmidrule(l{0.7em}){4-5} \cmidrule(l{0.7em}){6-7}
Method & Top 1$\uparrow$ & Top 5$\uparrow$ & Top 1$\uparrow$ & Top 5$\uparrow$ & Top 1$\uparrow$ &  Top 5$\uparrow$ \\

\midrule
SOTA (\cf~\cref{tab:somethingelse_eval,tab:cater_eval})  & 56.2 & 81.3 & 79.7 & 95.5 & 59.7 & 90.1\\
\model{} (Joint, Ours) & \textbf{61.1} & \textbf{85.4} & \textbf{80.6} & \textbf{97.2} & \textbf{73.7} & \textbf{95.6}\\
\bottomrule
\end{tabularx}
\vspace{-0.5 cm}
\end{table*}

\section{Surrogate Tasks and Overfitting}
\blue{To highlight that our surrogate tasks prevent our \model{} model from overfitting, we report training and test accuracy of our \model{} model trained with and without surrogate tasks on the compositional split of Something Else and on the moving camera split of CATER in \cref{tab:overfitting}.}

\begin{table*}[!t]
\centering
\small
\caption{Results highlighting that our \model{} model prevents overfitting.}
\vspace{-0.15cm}
\label{tab:overfitting}
\begin{tabularx}{\linewidth}{@{}Xcccc@{}}
\toprule
& \multicolumn{2}{c}{Something-Else (Top 1 $\uparrow$)} & \multicolumn{2}{c}{CATER: Moving Camera (Top 1 $\uparrow$)} \\
\cmidrule(l{0.7em}){2-3} \cmidrule(l{0.7em}){4-5}
Method & Train & Test & Train & Test \\
\midrule
\model{} (w/o Surrogate tasks) & 92.6 & 50.1 & 99.4	& 62.7\\
\model{} (Ours) & 87.3 & 62.0 & 89.7 & 80.4\\
\bottomrule
\end{tabularx}
\end{table*}

\blue{The results show that our surrogate tasks clearly prevent overfitting as the model is grounded to the fine-grained low-level details and is thus able to better ``understand'' the task at hand.}

\section{ACRE}
\label{apdx:acre}
\myparagraph{Additional training details.} 
We trained our \model{} model with the OPT-125M and OPT-1.3B backbone until convergence ($\sim 500$k iterations) with a batch size of 4. We use the AdamW optimizer \citep{LoshchilovH19} with a learning rate of $1 \times 10^{-5}$, $\,\,\,\beta_1 = 0.9, \, \beta_2 = 0.95 \,\,\,$ and $\lambda \,  \text{(weight decay)} = 0.1$ and gradient clipping with a norm of $1.0$.

\begin{table*}[h]
\small
\centering
\caption{Evaluation of our random prompting strategy on ACRE \cf~Table 2 in the main paper.}
\label{tab:acre_cot_eval}
\begin{tabularx}{\linewidth}{@{}X|c|c|c@{}}
\toprule
Method & Compositional $\uparrow$ & Systematic $\uparrow$ & Inference Speed (msec) $\uparrow$ \\
\midrule
ALOE \citep{abs-2012-08508} & 91.7 & 93.9 & - \\
\midrule
\model{} (Random 30\% during training, Ours) & 98.2 & 99.2 & \textbf{61} \\
\model{} (Every frame) & \textbf{99.3} & \textbf{99.5} & 1415 \\
\bottomrule
\end{tabularx}
\end{table*}

\myparagraph{Comparsion to ``Chain-of-thought''.} As described in Section 3.4 of the main paper, we prompt our LRR model to solve certain surrogate tasks at randomly selected time steps.
It is also possible to include surrogate tasks after every time-step. In this case, this would resemble a ``Chain of Thought'' \cite{wei2022chain} like process, where the final answer would depend upon the surrogate tasks solved at inference time.
However, this is very inefficient and thus impractical, especially for long video sequences.
We compare both of these approaches in \cref{tab:acre_cot_eval}. \model{} (Random 30\%, Ours) where we prompt our \model{} model with the OPT-125M backbone with a probability of 30\% an input video frame to solve surrogate tasks \emph{only} during training. \model{} (Every frame) where the model is prompted to solve surrogate tasks after every frame during training and inference. Note, that in case of the \model{} (Every frame) model it is necessary to solve surrogate tasks during inference as the (autoregressive LM) model always sees such tasks during training. We see from the results in \cref{tab:acre_cot_eval}, that the performance of \model{} (Random 30\% during training, Ours) is comparable with \model{} (Every frame) while the inference speed is order of magnitudes faster. We report the inference speed in milliseconds on a single Nvidia A100 GPU.
This is because our random prompting during training distills the relevant low-level information into the hidden states of the LM backbone at a feature level and thus does not require solving surrogate tasks at inference time in a COT like fashion.

\begin{figure*}[h]
\scriptsize
\centering
\begin{tabularx}{0.955\textwidth}{@{}ccccc|c@{}}
\toprule
\multicolumn{5}{c}{Context Trials} & Query \\ 
\midrule
\includegraphics[width=0.1155\linewidth]{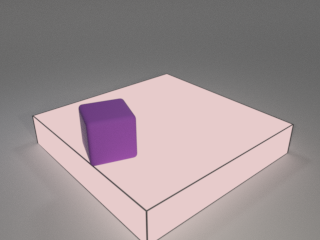} &
 &
\includegraphics[width=0.1155\linewidth]{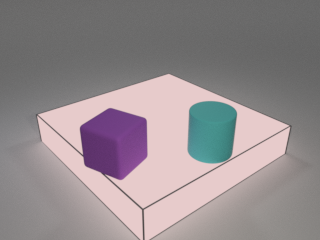} &
 &
\includegraphics[width=0.1155\linewidth]{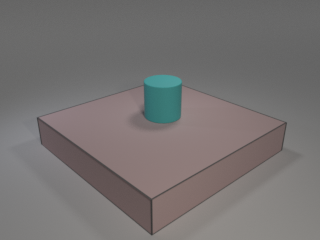} &
\includegraphics[width=0.1155\linewidth]{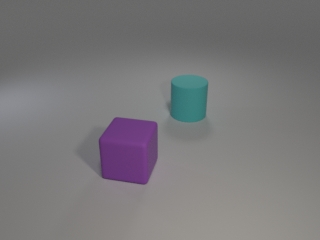} \\

\raggedright
\parbox{2.6cm}{<re-identify> \emph{ {\setulcolor{red} \ul{1,medium purple rubber cube}} } \\ <blicket> \emph{on.}} &
$\dots$ &
\parbox{2.6cm}{<re-identify>  \emph{{\setulcolor{red} \ul{1,medium purple rubber cube}}; \\ {\setulcolor{green} \ul{2, medium cyan rubber cylinder}}.} \\ <blicket> \emph{on.}} &
$\dots$ &
\parbox{2.6cm}{<re-identify>  \emph{{\setulcolor{green} \ul{2, medium cyan rubber cylinder}}. } \\ <blicket> \emph{off.}} &

\parbox{2.65cm}{<re-identify> \emph{{\setulcolor{red} \ul{1,medium purple rubber cube}}; {\setulcolor{green} \ul{2, medium cyan rubber cylinder}}} \\ <answer> \emph{yes.}}
\\

\midrule

\includegraphics[width=0.1155\linewidth]{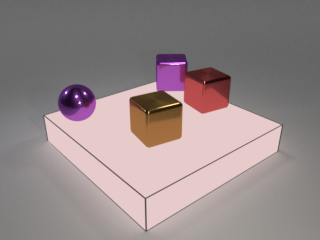} &
 &
\includegraphics[width=0.1155\linewidth]{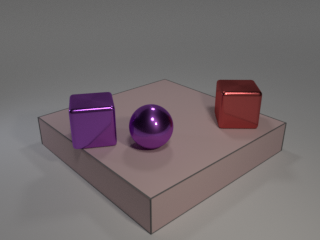} &
 &
\includegraphics[width=0.1155\linewidth]{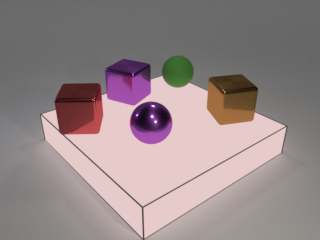} &
\includegraphics[width=0.1155\linewidth]{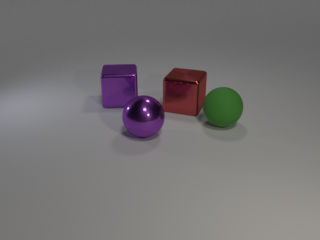} \\

\raggedright
\parbox{2.6cm}{<re-identify> \emph{ {\setulcolor{red} \ul{3, medium purple metal sphere}}; {\setulcolor{green} \ul{4,medium brown metal cube}}; {\setulcolor{blue} \ul{5,medium purple metal cube}}; {\setulcolor{cyan} \ul{6,medium red metal cube}}. } \\ <blicket> \emph{on.}} &
$\dots$ &
\parbox{2.6cm}{<re-identify>  \emph{{\setulcolor{blue} \ul{5, medium purple metal cube}}; {\setulcolor{red} \ul{3,medium purple metal sphere}}; {\setulcolor{cyan} \ul{6,medium red metal cube}}}. \\ <blicket> \emph{off.}} &
$\dots$ &
\parbox{2.6cm}{<re-identify>  \emph{ {\setulcolor{cyan} \ul{6, medium red metal cube}}; {\setulcolor{blue} \ul{5,medium purple metal cube}}; {\setulcolor{red} \ul{3,medium purple metal sphere}}; {\setulcolor{magenta} \ul{7,medium green rubber sphere}}; {\setulcolor{green} \ul{4,medium brown metal cube}}. } \\ <blicket> \emph{on.}} &

\parbox{2.65cm}{<re-identify> \emph{{\setulcolor{blue} \ul{5, medium purple metal cube}}; {\setulcolor{red} \ul{3, medium purple metal sphere}}; {\setulcolor{cyan} \ul{6, medium red metal cube}};  {\setulcolor{magenta} \ul{7, medium green rubber sphere}}.} \\ <answer> \emph{maybe.}}
\\

\midrule

\includegraphics[width=0.1155\linewidth]{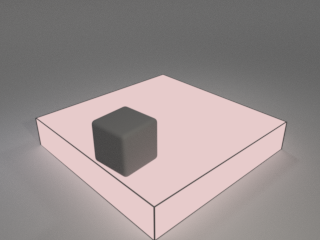} &
 &
\includegraphics[width=0.1155\linewidth]{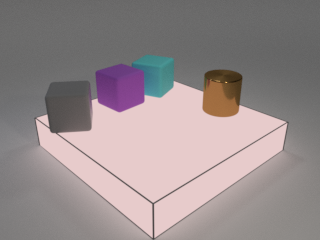} &
 &
\includegraphics[width=0.1155\linewidth]{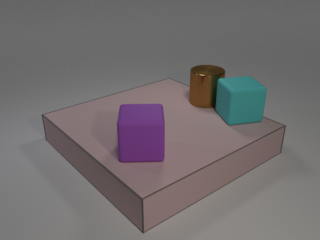} &
\includegraphics[width=0.1155\linewidth]{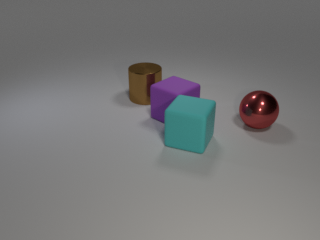} \\

\raggedright
\parbox{2.6cm}{<re-identify> \emph{ {\setulcolor{red} \ul{3, medium gray rubber cube}}. } \\ <blicket> \emph{on.}} &
$\dots$ &
\parbox{2.6cm}{<re-identify>  \emph{ {\setulcolor{red} \ul{3, medium gray rubber cube}}; {\setulcolor{green} \ul{4, medium purple rubber cube}}; {\setulcolor{blue} \ul{5, medium cyan rubber cube}}; {\setulcolor{cyan} \ul{6,medium brown metal cylinder}} }. \\ <blicket> \emph{on.}} &
$\dots$ &
\parbox{2.6cm}{<re-identify>  \emph{ {\setulcolor{green} \ul{4, medium purple rubber cube}}; {\setulcolor{cyan} \ul{6,medium brown metal cylinder}}; {\setulcolor{blue} \ul{5, medium cyan rubber cube}}. } \\ <blicket> \emph{off.}} &

\parbox{2.65cm}{<re-identify> \emph{ {\setulcolor{blue} \ul{5, medium cyan rubber cube}}; {\setulcolor{red} \ul{3, medium gray rubber cube}}; {\setulcolor{green} \ul{4, medium purple rubber cube}}; {\setulcolor{cyan} \ul{6, medium brown metal cylinder}}.} \\ <answer> \emph{yes.}}
\\

\bottomrule
\end{tabularx}
\caption{Example solutions to surrogate tasks generated by our \model{} model on ACRE. Re-identified objects across context trials are underlined in the same color. }
\label{tab:acre_qualitative_supp}
\end{figure*}

\myparagraph{Additional qualitative examples.} We include additional qualitative examples in \cref{tab:acre_qualitative_supp} highlighting surrogate tasks. These examples illustrate that using our surrogate re-identification task, our \model{} model can aggregate information from multiple context trials to arrive at the final answer -- following the paradigm of ``Look, Remember, Reason''.

\section{Something-Else}
\myparagraph{Additional training details.} 
We trained our \model{} model with the OPT-125M and OPT-1.3B backbone until convergence ($\sim 700$k iterations) with a batch size of 4. We use the AdamW optimizer \citep{LoshchilovH19} with a learning rate of $1 \times 10^{-5}$, $\,\,\,\beta_1 = 0.9, \, \beta_2 = 0.95 \,\,\,$ and $\lambda \,  \text{(weight decay)} = 0.1$ and gradient clipping with a norm of $1.0$. We also employ random data augmentation using RandAugment \citep{CubukZSL20} with a magnitude of 15 and label smoothing with $\epsilon = 0.3$, where applicable (for fairness).

\begin{figure*}[t]
\small
\centering
\begin{tabular}{cccc}
\toprule
\includegraphics[width=0.24\linewidth]{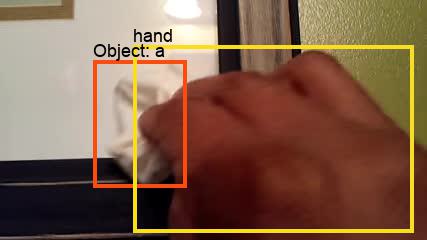} & 
\includegraphics[width=0.24\linewidth]{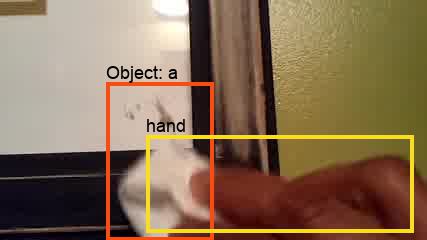} &
\includegraphics[width=0.24\linewidth]{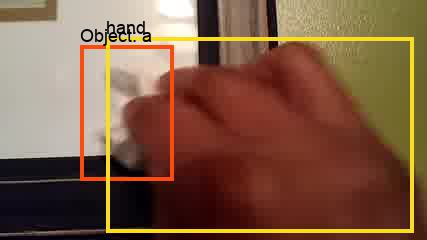} &
\includegraphics[width=0.24\linewidth]{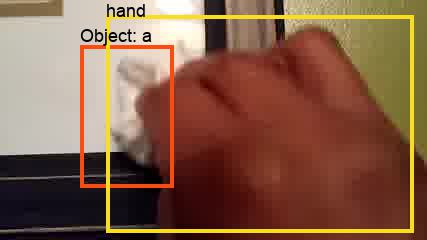} \\
\multicolumn{4}{c}{Answer: Pretending or failing to wipe [something] off of [something]. } \\
\midrule
\includegraphics[width=0.24\linewidth]{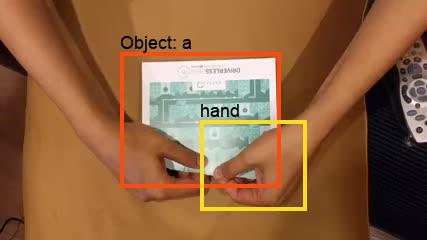} & 
\includegraphics[width=0.24\linewidth]{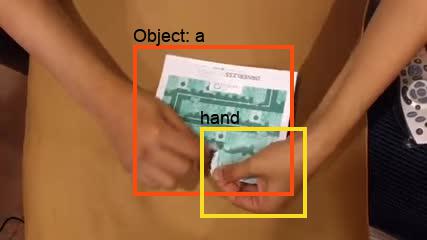} &
\includegraphics[width=0.24\linewidth]{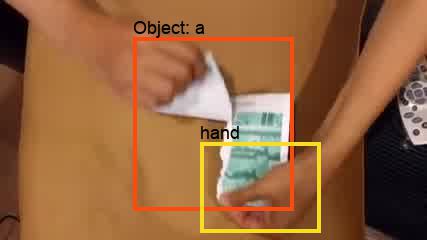} &
\includegraphics[width=0.24\linewidth]{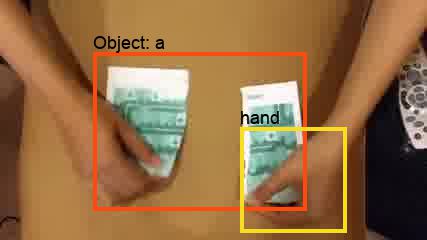} \\
\multicolumn{4}{c}{Answer: Tearing [something] into two pieces. } \\
\midrule
\includegraphics[width=0.24\linewidth]{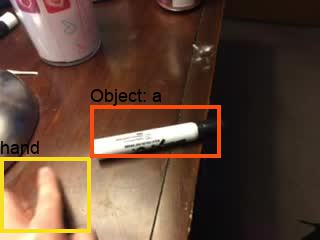} & 
\includegraphics[width=0.24\linewidth]{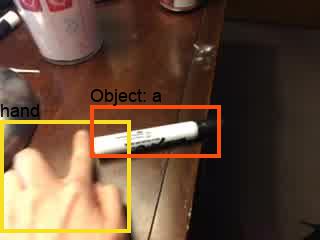} &
\includegraphics[width=0.24\linewidth]{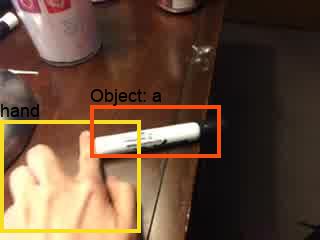} &
\includegraphics[width=0.24\linewidth]{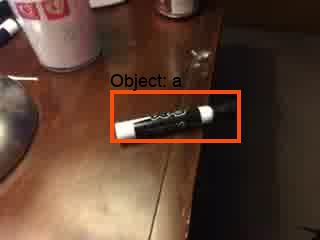} \\
\multicolumn{4}{c}{Answer: Pushing [something] so that it almost falls off but doesn’t. } \\
\midrule
\includegraphics[width=0.24\linewidth]{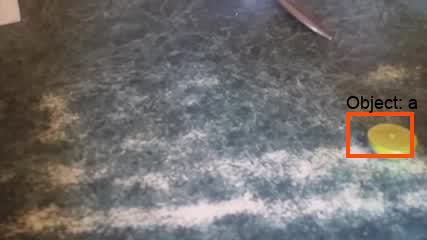} & 
\includegraphics[width=0.24\linewidth]{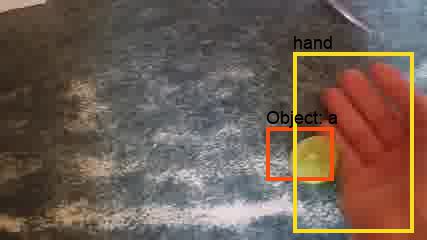} &
\includegraphics[width=0.24\linewidth]{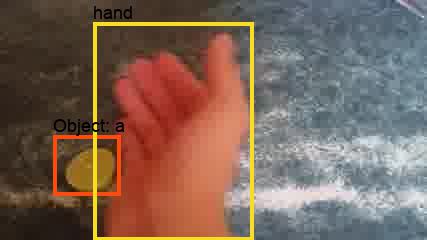} &
\includegraphics[width=0.24\linewidth]{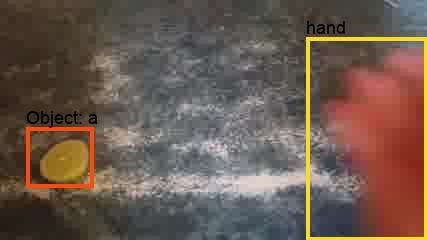} \\
\multicolumn{4}{c}{Answer: Pushing [something] from right to left. } \\
\midrule
\includegraphics[width=0.24\linewidth]{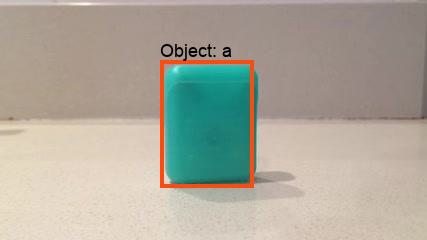} & 
\includegraphics[width=0.24\linewidth]{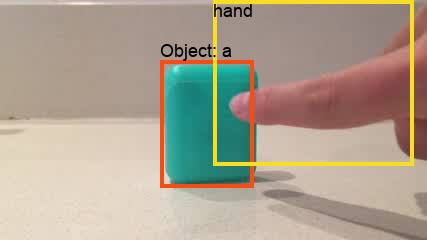} &
\includegraphics[width=0.24\linewidth]{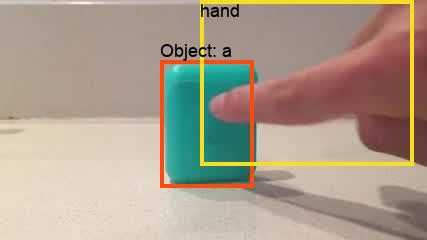} &
\includegraphics[width=0.24\linewidth]{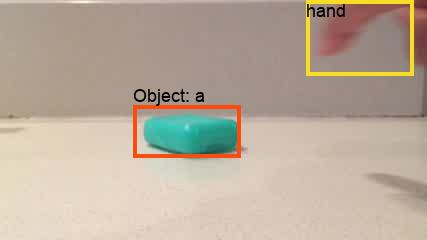} \\
\multicolumn{4}{c}{Answer: Poking [something] so that it falls over. } \\
\midrule
\includegraphics[width=0.24\linewidth]{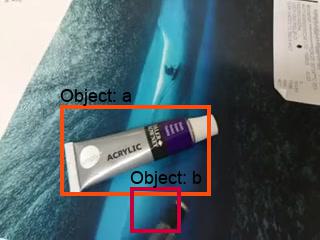} & 
\includegraphics[width=0.24\linewidth]{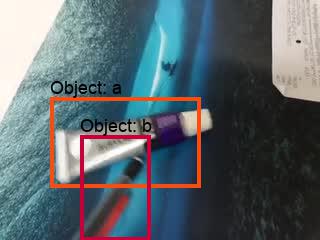} &
\includegraphics[width=0.24\linewidth]{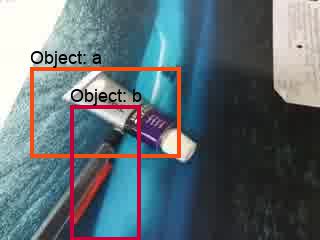} &
\includegraphics[width=0.24\linewidth]{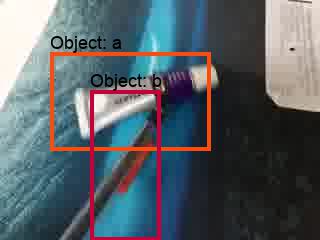} \\
\multicolumn{4}{c}{Answer: Pushing [something] with [something]. } \\
\midrule
\includegraphics[width=0.24\linewidth]{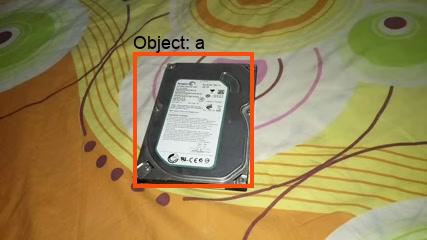} & 
\includegraphics[width=0.24\linewidth]{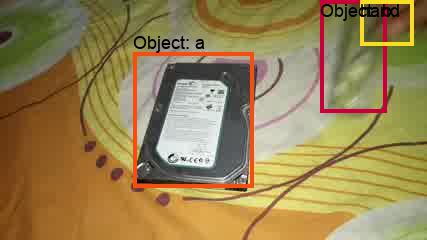} &
\includegraphics[width=0.24\linewidth]{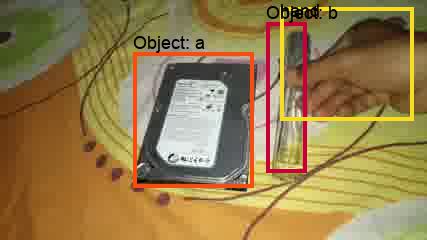} &
\includegraphics[width=0.24\linewidth]{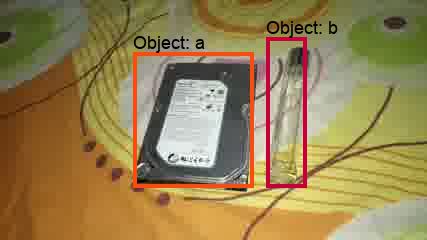} \\
\multicolumn{4}{c}{Answer: Putting [something] next to [something]. } \\
\bottomrule
\end{tabular}
\caption{Example solutions to surrogate tasks generated by our \model{} model on Something-Else, highlighting object tracks and bounding boxes}
\label{tab:somethingelse_qualitative_supp}
\end{figure*}

\myparagraph{Additional qualitative examples.} \blue{The qualitative examples in \cref{tab:somethingelse_qualitative} show that our \model{} model can deal with visually complex real-world scenarios such as severe occlusions, \eg, the silver ball in \cref{tab:somethingelse_qualitative} (bottom row) is occluded by the hand. Here in \cref{tab:somethingelse_qualitative_supp}, we include additional qualitative examples further highlighting that our \model{} model can deal with visually complex real-world scenarios such as severe deformations, \eg, the piece of paper in \cref{tab:somethingelse_qualitative_supp} second row which is torn into two pieces; severe changes in appearance, \eg, the pen in \cref{tab:somethingelse_qualitative_supp} third row, the teal case in the fifth row; motion blur in \cref{tab:somethingelse_qualitative_supp} last row.}

Finally, these examples illustrate that using our surrogate tracking task, our \model{} model can aggregate information from multiple context trials to arrive at the final answer -- following the paradigm of ``Look, Remember, Reason''.

\section{CATER}
\label{apdx:cater}
\myparagraph{Additional training details.} 
We trained our \model{} model with the OPT-125M backbone until convergence ($\sim 600$k iterations) with a batch size of 4. We use the AdamW optimizer \citep{LoshchilovH19} with a learning rate of $1 \times 10^{-5}$, $\,\,\,\beta_1 = 0.9, \, \beta_2 = 0.95 \,\,\,$ and $\lambda \,  \text{(weight decay)} = 0.1$. We use gradient clipping with a norm of $1.0$.

\begin{figure*}[!t]
\small
\centering
\begin{tabularx}{0.98\linewidth}{@{}cccc@{}}
\toprule

\includegraphics[width=0.2\linewidth]{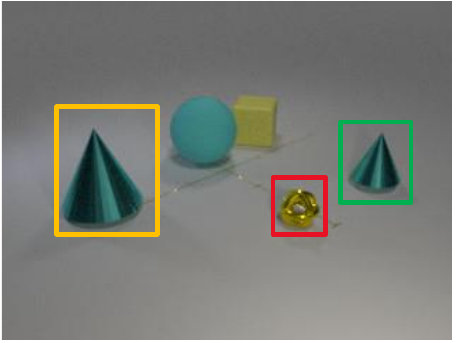} & 
\includegraphics[width=0.2\linewidth]{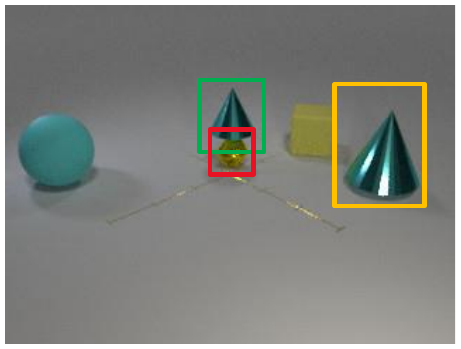} &
\includegraphics[width=0.2\linewidth]{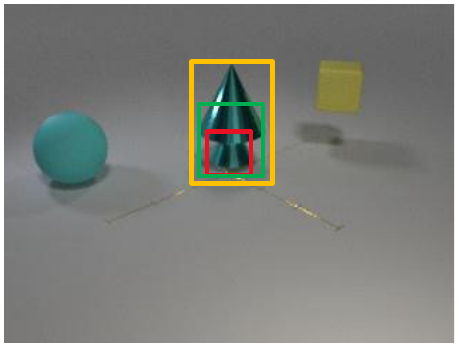} &
\includegraphics[width=0.2\linewidth]{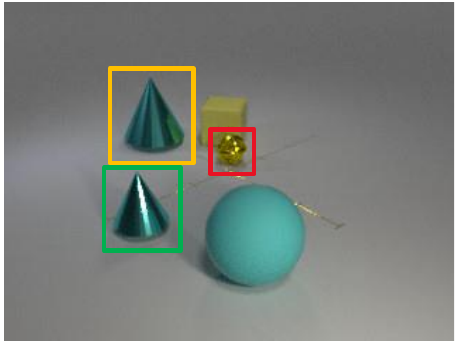}\\
\setulcolor{amber}  \ul{2},{\setulcolor{green} \ul{29}},{\setulcolor{red} \ul{17}} &
\setulcolor{amber}  \ul{35},{\setulcolor{green} \ul{20}},{\setulcolor{red} \ul{20}} & 
\setulcolor{amber}  \ul{20},{\setulcolor{green} \ul{20}},{\setulcolor{red} \ul{20}} & 
\setulcolor{amber}  \ul{12},{\setulcolor{green} \ul{3}},{\setulcolor{red} \ul{20}}\\
\midrule

\includegraphics[width=0.227\linewidth]{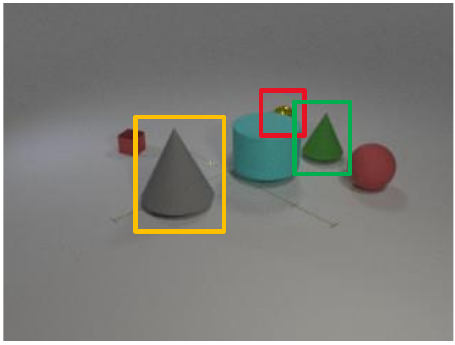} & 
\includegraphics[width=0.227\linewidth]{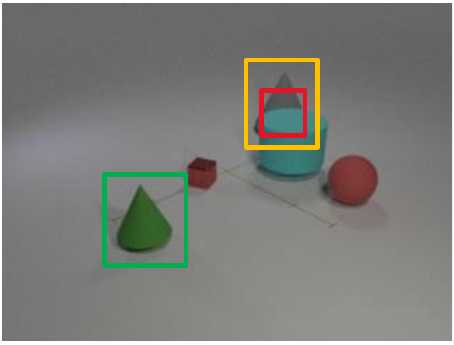} &
\includegraphics[width=0.227\linewidth]{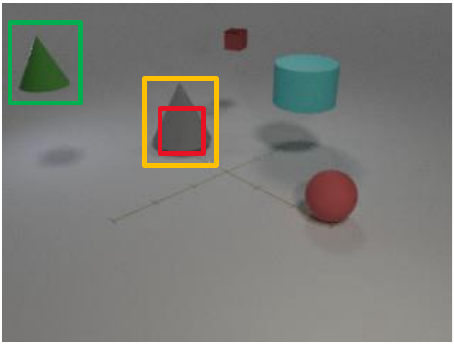} &
\includegraphics[width=0.227\linewidth]{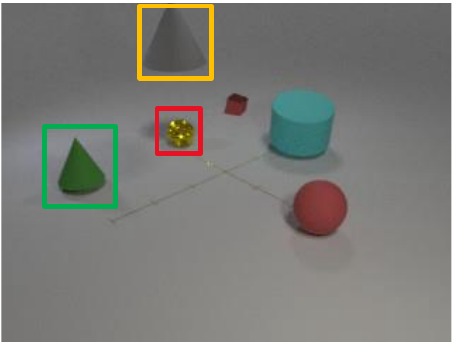}\\
\setulcolor{amber}  \ul{9},{\setulcolor{green} \ul{34}},{\setulcolor{red} \ul{32}} &
\setulcolor{amber}  \ul{32},{\setulcolor{green} \ul{3}},{\setulcolor{red} \ul{32}} & 
\setulcolor{amber}  \ul{19},{\setulcolor{green} \ul{0}},{\setulcolor{red} \ul{19}} & 
\setulcolor{amber}  \ul{7},{\setulcolor{green} \ul{2}},{\setulcolor{red} \ul{19}}\\
\midrule

\includegraphics[width=0.227\linewidth]{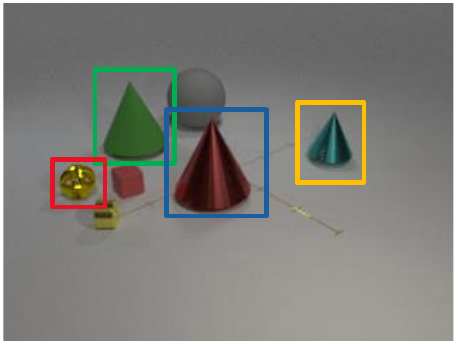} & 
\includegraphics[width=0.227\linewidth]{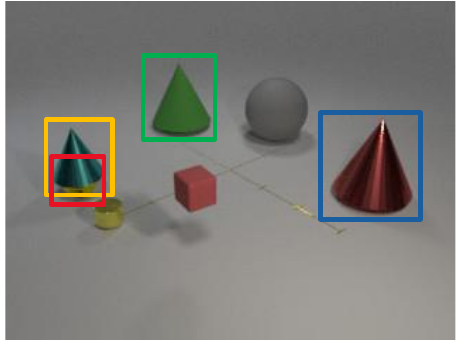} &
\includegraphics[width=0.227\linewidth]{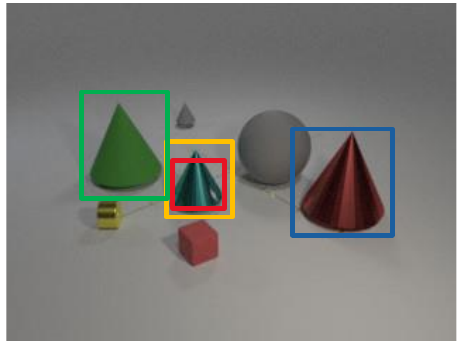} &
\includegraphics[width=0.227\linewidth]{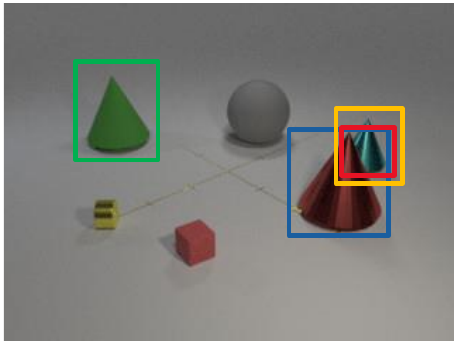}\\
\setulcolor{amber}  \ul{28},{\setulcolor{green} \ul{12}},{\setulcolor{blue} \ul{15}},{\setulcolor{red} \ul{1}} &
\setulcolor{amber}  \ul{1},{\setulcolor{green} \ul{18}},{\setulcolor{blue} \ul{29}},{\setulcolor{red} \ul{1}} &
\setulcolor{amber}  \ul{9},{\setulcolor{green} \ul{7}},{\setulcolor{blue} \ul{23}},{\setulcolor{red} \ul{9}} &
\setulcolor{amber}  \ul{34},{\setulcolor{green} \ul{6}},{\setulcolor{blue} \ul{23}},{\setulcolor{red} \ul{34}}\\

\midrule

\includegraphics[width=0.227\linewidth]{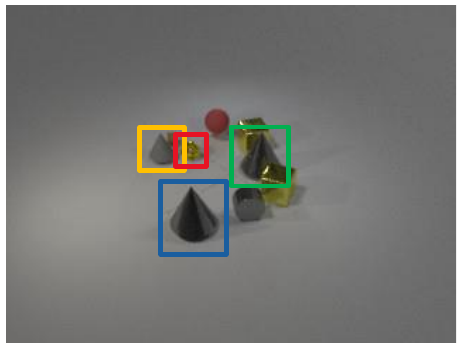} & 
\includegraphics[width=0.227\linewidth]{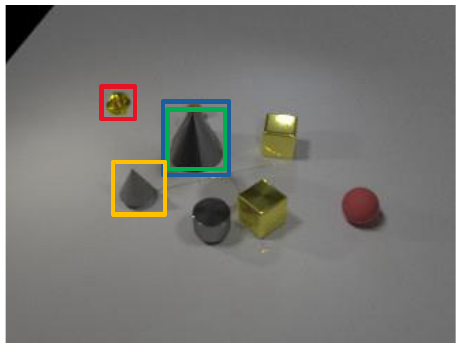} &
\includegraphics[width=0.227\linewidth]{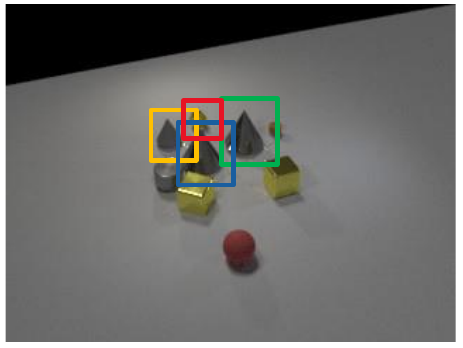} &
\includegraphics[width=0.227\linewidth]{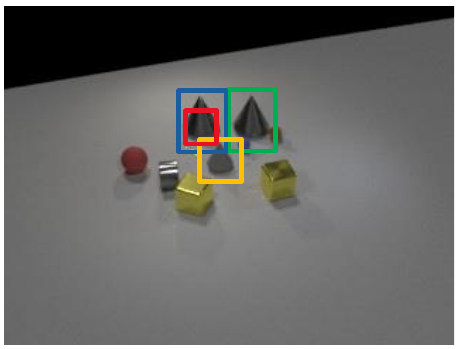}\\
\setulcolor{amber}  \ul{4},{\setulcolor{green} \ul{20}},{\setulcolor{blue} \ul{29}},{\setulcolor{red} \ul{9}} &
\setulcolor{amber}  \ul{23},{\setulcolor{green} \ul{9}},{\setulcolor{blue} \ul{22}},{\setulcolor{red} \ul{11}} &
\setulcolor{amber}  \ul{23},{\setulcolor{green} \ul{9}},{\setulcolor{blue} \ul{22}},{\setulcolor{red} \ul{11}} &
\setulcolor{amber}  \ul{15},{\setulcolor{green} \ul{4}},{\setulcolor{blue} \ul{11}},{\setulcolor{red} \ul{11}}\\

\midrule

\includegraphics[width=0.227\linewidth]{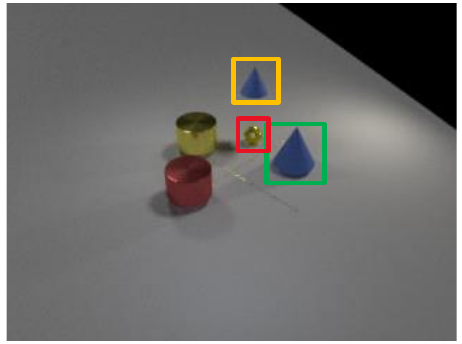} & 
\includegraphics[width=0.227\linewidth]{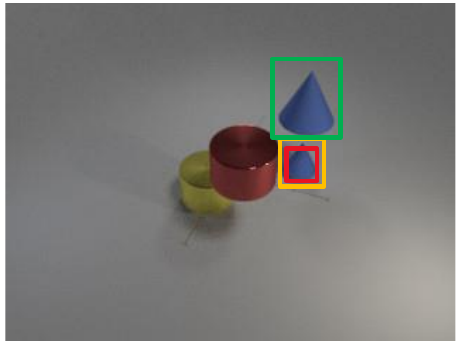} &
\includegraphics[width=0.227\linewidth]{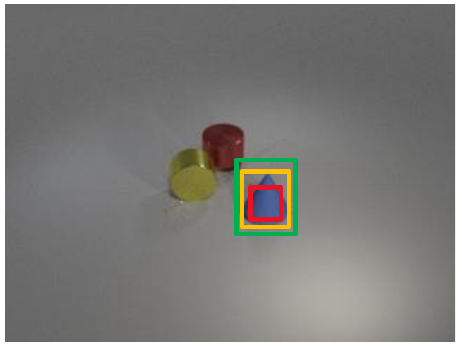} &
\includegraphics[width=0.227\linewidth]{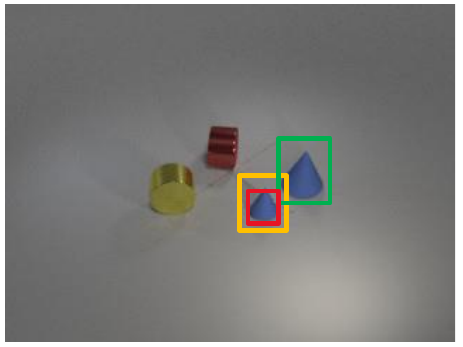}\\
\setulcolor{amber}  \ul{22},{\setulcolor{green} \ul{10}},{\setulcolor{red} \ul{22}} &
\setulcolor{amber}  \ul{22},{\setulcolor{green} \ul{22}},{\setulcolor{red} \ul{22}} & 
\setulcolor{amber}  \ul{17},{\setulcolor{green} \ul{17}},{\setulcolor{red} \ul{17}} & 
\setulcolor{amber}  \ul{17},{\setulcolor{green} \ul{29}},{\setulcolor{red} \ul{17}}\\

\bottomrule
\end{tabularx}
\caption{Example solution to the tracking surrogate tasks generated by our \model{} model on CATER.  We show the predicted grid locations of the cones and the snitch below.}
\label{tab:cater_qualitative_supp}
\end{figure*}

\myparagraph{Additional qualitative examples.} We include additional qualitative examples including both the static and moving camera splits in \cref{tab:cater_qualitative_supp}, highlighting the surrogate tracking task. We see that our \model{} model can successfully deal with containment in both static camera (rows 1-3; \cref{tab:cater_qualitative_supp}) and moving camera (row 4-5; \cref{tab:cater_qualitative_supp}) settings, due to our rationale that explicitly tracks cones and the snitch. Note that the example in row 5 in \cref{tab:cater_qualitative_supp} from the moving camera split is especially challenging due to recursive containment. 

\end{document}